\documentclass[lettersize,journal]{IEEEtran}
\usepackage{amsmath,amsfonts}
\usepackage{algorithmic}
\usepackage{algorithm}
\usepackage{array}
\usepackage[caption=false,font=normalsize,labelfont=sf,textfont=sf]{subfig}
\usepackage{textcomp}
\usepackage{stfloats}
\usepackage{url}
\usepackage{verbatim}
\usepackage{graphicx}
\usepackage{cite}
\usepackage{hyperref}
\usepackage{booktabs}
\usepackage{makecell}
\usepackage{multirow}
\usepackage{colortbl}
\usepackage[table]{xcolor}
\hypersetup{
    colorlinks=true,   % 启用颜色
    linkcolor=blue,    % 设置内部链接（如章节链接）的颜色
    citecolor=blue,     % 设置文献引用的颜色
    filecolor=magenta, % 设置文件链接的颜色
    urlcolor=blue      % 设置网址链接的颜色
}

\definecolor{venuedark}{rgb}{0.7, 0.7, 0.7}

\newcommand{\rvs}[1]{{#1}}
\newcommand{\rrvs}[1]{{#1}}

\begin{document}

\title{Current Injection Spiking Neural Network for Infrared and Visible Image Fusion}

\author{
        Rui Zhao,~\IEEEmembership{Member,~IEEE,}
        Zhuoyuan Li,
        Wenrui Li,~\IEEEmembership{Member,~IEEE,}
        Yanchen Dong,
        Yajing Zheng,~\IEEEmembership{Member,~IEEE,}\\
        Giuseppe Valenzise,~\IEEEmembership{Senior Member,~IEEE,}
        Weisi Lin,~\IEEEmembership{Fellow,~IEEE}
        % <-this % stops a space
\thanks{\textit{Corresponding author: Weisi Lin.}}
\thanks{Rui Zhao and Weisi Lin are with the College of Computing and Data Science, Nanyang Technological University, Singapore (E-mail: ruizhao26@gmail.com, wslin@ntu.edu.sg).}
\thanks{Zhuoyuan Li is with the Department of Electrical and Electronic Engineering, The Hong Kong Polytechnic University, Hong Kong (E-mail: zhuoyuan1997@gmail.com).}
\thanks{Wenrui Li is with the Department of Computer Science and Technology, Harbin Institute of Technology, Harbin, China. (e-mail: liwr@hit.edu.cn).}
\thanks{Yanchen Dong and Yajing Zheng are with the State Key Laboratory for Multimedia Information Processing, School of Computer Science, Peking University, Beijing, China (E-mail: yanchendong@stu.pku.edu.cn, yj.zheng@pku.edu.cn).}
\thanks{Giuseppe Valenzise is with the Université Paris-Saclay, CentraleSupélec, CNRS, Laboratoire des Signaux et Systémes, Gif-sur-Yvette, France (E-mail: giuseppe.valenzise@l2s.centralesupelec.fr).}
}



\maketitle

\begin{abstract}
% Infrared and visible image fusion (IVIF) aims to integrate the complementary information of two modalities into a single image with richer scene content. While existing methods are largely built on artificial neural networks (ANNs), spiking neural networks (SNNs) offer a fundamentally different computation paradigm. Their bio-inspired dynamics, in which a neuron accumulates input currents at its membrane potential before firing, provide a natural substrate for integrating information from two sources. Motivated by this, we propose CIS-Fuse, a spiking neural network for IVIF that performs cross-modal fusion directly at the membrane-potential level. 
% While existing methods are largely built on artificial neural networks (ANNs), spiking neural networks (SNNs) offer a promising route to sparse and energy-efficient fusion computation. 
\rvs{Infrared and visible image fusion (IVIF) aims to integrate the complementary information of two modalities into a single image with richer scene content. 
\rrvs{While existing methods are largely built on artificial neural networks (ANNs), which densely compute over all activations, spiking neural networks (SNNs) communicate through sparse binary spikes and perform computation only where and when a spike occurs, offering a route to more energy-efficient fusion.}
However, directly applying SNNs to IVIF creates a fundamental tension: cross-modal fusion relies on fine-grained responses from both modalities, whereas communicating only through binary spikes can discard complementary cues that remain below the firing threshold. The membrane potential retains these subthreshold responses before firing, letting both modalities jointly shape the output when integrated at this stage. Building on this, we propose CIS-Fuse, a spiking neural network that performs cross-modal fusion directly at the membrane-potential level.}
At its core is the current injection spiking (CIS) operator, which injects one modality as a gated auxiliary current into the driving neuron of the other, so the two integrate before spike firing. A per-channel learnable injection strength further allows different feature channels to adaptively regulate the modulation magnitude. Building on CIS, we construct a bidirectional cross-modal fusion (BCMF) module and deploy it on a dual-branch architecture with asymmetric stacking depths, where the two branches develop a clear functional specialization. 
% A reparameterized output head and an output-mean supervision scheme are further introduced to balance training expressiveness and inference efficiency. 
Extensive experiments on four IVIF benchmarks and on downstream object detection and semantic segmentation tasks show that CIS-Fuse achieves fusion quality on par with state-of-the-art ANN-based methods, \rrvs{while inheriting the energy efficiency of spike-based computation, with roughly an order of magnitude lower inference energy than the similarly-sized ANN-based DCEvo.} Code will be released upon publication. 
\end{abstract}


\section{Introduction}
\IEEEPARstart{I}{nfrared} and visible image fusion (IVIF)~\cite{toet1989merging,liu2024infrared} aims to integrate the complementary information of two modalities into a single image with richer scene content. Visible images provide rich texture and color details that align well with human visual perception, but they degrade severely under low illumination, smoke, or nighttime conditions. Infrared images, in contrast, sense thermal radiation and remain robust in such adverse conditions, reliably highlighting thermal targets such as pedestrians and vehicles. The physical complementarity of the two modalities makes the fused images broadly valuable for downstream visual tasks, including autonomous driving~\cite{ha2017mfnet}, object detection~\cite{tan2024rle} and tracking~\cite{li2018cross}, and semantic segmentation~\cite{li2024object,tang2022image}.

Recent advances in deep learning have substantially propelled IVIF research, driving a transition from hand-crafted fusion rules toward fully learnable architectures and, more recently, from purely visual quality optimization toward joint optimization with high-level perception. Early CNN-based methods~\cite{xu2020u2fusion,zhang2021sdnet,zhao2023cddfuse} learn modality-shared and modality-specific representations end-to-end, while Transformer-based methods~\cite{yi2024text,zhu2024task} broaden the receptive field to better capture cross-modal long-range dependencies. In parallel, generative formulations based on adversarial~\cite{wang2022unsupervised,ma2019fusiongan} or diffusion~\cite{zhao2023ddfm,yue2023dif} models pursue more realistic fused appearance. Downstream-task-driven frameworks~\cite{liu2022target,liu2023multi,liu2025dcevo} couple fusion with detection or segmentation so that the fused image directly serves perception.

\begin{figure}
    \centering
    \includegraphics[width=\linewidth]{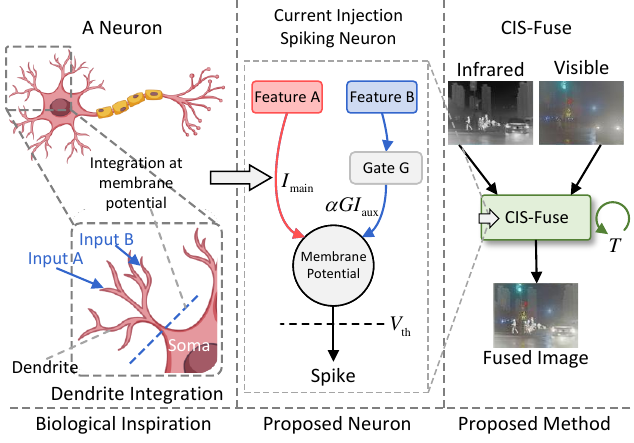}
    \vspace*{-15 pt}
    \caption{Biological inspiration and design philosophy of CIS-Fuse. \textit{Left}: Dendrites receive inputs from multiple sources and integrate them at the membrane potential before the soma fires a spike, known as dendritic integration. \textit{Middle}: The proposed CIS neuron treats one modality as the main driving current $\mathbf{I}_\text{main}$ and the other as a gated auxiliary current $\alpha G \mathbf{I}_\text{aux}$, enabling cross-modal integration before spike firing. \textit{Right}: CIS-Fuse processes infrared and visible inputs through $T$ recurrent spiking steps to produce the fused image.}
    \label{fig:teaser}
\vspace*{-10pt}
\end{figure}

% Despite these advances, the above methods are largely built on the artificial neural network (ANN) paradigm, where information flows as continuous real-valued activations through dense matrix operations. Spiking neural networks (SNNs)~\cite{maass1997networks,tavanaei2019deep} instead build on a different computation model that mimics biological neural processing: neurons accumulate input currents into a membrane potential and fire binary spikes only when that potential crosses a threshold. Beyond the event-driven sparsity that enables low-power computation on neuromorphic hardware~\cite{roy2019towards}, a property more fundamental to our purpose is that each neuron continuously accumulates and integrates its input currents at the membrane potential before firing, providing a natural substrate for aggregating information from multiple sources~\cite{eshraghian2023training}.

Despite these advances, the above methods are largely built on the artificial neural network (ANN) paradigm, where information is represented by continuous real-valued activations and processed through dense multiply-accumulate operations. \rvs{
% The dense computation in ANNs can be costly for resource-constrained perception systems in which infrared and visible sensing needs to operate continuously. 
\rrvs{The dense computation in ANNs can be costly for resource-constrained perception systems, such as onboard drones or mobile robots, in which infrared and visible sensing needs to operate continuously.}
Spiking neural networks (SNNs)~\cite{maass1997networks,tavanaei2019deep}, in contrast, communicate through sparse binary spikes and perform event-driven computation, offering a promising route to energy-efficient visual processing on neuromorphic hardware~\cite{roy2019towards}. These characteristics make SNNs an appealing computational paradigm for image fusion~\cite{li2025multi}.}
\rrvs{SNNs have been widely employed to process neuromorphic data~\cite{zhu2022event,zheng2026spikecv}, such as the output of event-based cameras~\cite{li2025brain, hagenaars2021self}, owing to their asynchronous nature. They have also been applied to conventional image-processing tasks to reduce energy consumption~\cite{luo2024integer,xiao2026spiking}. In this work, we adopt the latter perspective and leverage SNNs for energy-efficient image fusion of conventional infrared and visible images.}

% This integrative nature makes the membrane potential an appealing yet underexplored substrate for IVIF, whose essence is precisely to aggregate complementary cues from two modalities. However, transplanting SNNs into cross-modal fusion is not a trivial paradigm substitution: a naive port that lets the two modalities meet only at the binary spikes squanders their continuous responses, discarding the fine-grained modulation on which fusion relies. The distinctive computational asset of SNNs for fusion thus lies not in the spike itself, but in the continuous membrane potential integration that precedes it. Inspired by dendritic integration in biological neurons~\cite{london2005dendritic}, where a neuron integrates input currents from multiple sources at the membrane potential before firing, we argue that cross-modal fusion should occur at this pre-spike integration stage, where genuine multi-source integration takes place.
\rvs{However, applying SNNs to IVIF cannot simply replace ANN activations with binary spikes. Cross-modal fusion relies on fine-grained responses from both modalities, yet a weak response below the firing threshold may still carry complementary evidence when combined with the other modality. If cross-modal interaction occurs only after spike firing, such subthreshold responses are discarded before they can contribute to fusion. This creates a fundamental tension between the binary communication of SNNs and the fine-grained information integration required by IVIF.}

% Building on this insight, we introduce the Current Injection Spiking (CIS) operator. CIS injects the auxiliary modality as a gated, content-dependent current into the main modality neuron's membrane potential, so that the two modalities integrate before the spike is emitted, while a per-channel learnable strength $\alpha$ lets each feature channel adaptively regulate its own modulation magnitude. On top of CIS, we further construct a Bidirectional Cross-Modal Fusion (BCMF) module that injects in both directions in an alternating manner, letting the IR and VIS representations mutually refine each other before firing.

\rvs{The membrane potential provides a natural way to resolve this tension. Before spike firing, it retains and accumulates input currents, including responses that remain below the firing threshold. Integrating the two modalities at this pre-spike stage therefore allows their complementary evidence to jointly determine neuronal firing, while subsequent information propagation remains spike-based. From this perspective, membrane-potential-level fusion reconciles fine-grained cross-modal integration with sparse spike-driven computation, echoing how a biological neuron combines currents from multiple sources before firing~\cite{london2005dendritic}.}

\rvs{To realize this pre-spike fusion principle, we introduce the Current Injection Spiking (CIS) operator.} CIS injects the auxiliary modality as a gated, content-dependent current into the main modality neuron's membrane potential, so that the two modalities integrate before the spike is emitted, while a per-channel learnable strength $\alpha$ lets each feature channel adaptively regulate its own modulation magnitude. On top of CIS, we further construct a Bidirectional Cross-Modal Fusion (BCMF) module that injects in both directions in an alternating manner, letting the IR and VIS representations mutually refine each other before firing.

We deploy BCMF on a dual-branch architecture with asymmetric stacking depths, imposing only a soft, non-binding bias at initialization. After training, the per-channel $\alpha$ distributions of the two branches settle into well-separated regimes, indicating that the network arrives at a clear functional specialization between the two branches~\cite{livingstone1988segregation}. Finally, we introduce a reparameterized output head for stronger reconstruction and a $T$-step output-mean supervision that aligns training with the aggregated multi-step output.

We name the proposed method \textbf{CIS-Fuse}. Extensive experiments on four IVIF benchmarks, namely RoadScene~\cite{xu2020fusiondn}, MSRS~\cite{tang2022piafusion}, M3FD~\cite{liu2022target}, and FMB~\cite{liu2023multi}, together with downstream object detection and semantic segmentation, show that CIS-Fuse attains fusion quality on par with state-of-the-art ANN-based methods while preserving the energy efficiency of spike-based computation. The main contributions of this paper are summarized as follows.

\begin{itemize}
    % \item[(1)] We revisit IVIF from the perspective of spiking neural networks and propose to perform cross-modal fusion at the continuous membrane-potential level, rather than after binary spike firing, bringing the integrative dynamics of SNNs into the fusion task.
    % \item[(1)] \rvs{We explore IVIF under the SNN paradigm and propose to perform cross-modal fusion at the membrane-potential level before spike firing, helping preserve fine-grained cross-modal responses during spiking computation.}
    \item[(1)] \rrvs{We approach IVIF from the perspective of where fusion should take place within the spiking dynamics, and propose to integrate the two modalities at the continuous membrane potential before spike firing, so that fine-grained cross-modal responses are preserved rather than lost at the binary spike.}
    \item[(2)] We propose the CIS operator, which injects one modality as a gated, per-channel-scaled current into the membrane potential of the other, together with its bidirectional extension BCMF. Deployed on a dual-branch architecture with asymmetric stacking depths, CIS induces an emergent functional specialization between the two branches.
    \item[(3)] Extensive experiments on four IVIF benchmarks and on downstream detection and segmentation show that CIS-Fuse matches state-of-the-art ANN-based methods in fusion quality while retaining the energy efficiency of spike-based computation.
\end{itemize}

% \begin{itemize}
%     \item[(1)] We propose to perform cross-modal integration at the membrane potential level for SNNs, bringing the continuous neural dynamics of the SNNs into the IVIF task.
%     \item[(2)] We propose the CIS operator and its bidirectional extension BCMF, which perform cross-modal fusion by injecting one modality as a modulating current into the membrane potential of the other. We also deployed on a dual-branch architecture with asymmetric stacking depths that induces functional divergence.
%     \item[(3)] We integrate RepHead reparameterization with output-mean supervision to construct the end-to-end CIS-Fuse framework. Experiments on multiple IVIF benchmarks and downstream tasks validate the effectiveness of the proposed method.
% \end{itemize}

% \noindent(1) We propose a paradigm of performing cross-modal integration at the membrane potential level, bringing the continuous neural dynamics that are unique to SNNs into the IVIF task.

% \noindent(2) We propose the CIS operator and its bidirectional extension BCMF, which inject one modality as a modulating current into the membrane potential of another, and deploy them on a dual-branch architecture with asymmetric stacking depths that induces parameter-level functional divergence.

% \noindent(3) We integrate RepHead reparameterization with output-mean supervision to construct the end-to-end CIS-Fuse framework, and extensive experiments on multiple IVIF benchmarks and downstream tasks validate the effectiveness of the proposed method.

\section{Related Work}

\subsection{Infrared and Visible Image Fusion}

\noindent\textbf{Fusion for visual enhancement.}

Reconstruction-oriented IVIF methods aim to improve fused-image quality. 
CNN-based methods commonly design feature decomposition and aggregation strategies to capture complementary cross-modal information, e.g., U2Fusion~\cite{xu2020u2fusion} adaptively preserves informative source content, SDNet~\cite{zhang2021sdnet} formulates fusion as gradient and intensity reconstruction, and CDDFuse~\cite{zhao2023cddfuse} decomposes features into base and detail components. Others introduce lightweight or high-order designs, such as DeFusion~\cite{liang2022fusion}, LRRNet~\cite{li2023lrrnet}, and SHIP~\cite{zheng2024probing}.
% CNN-based methods commonly design feature decomposition and aggregation strategies to capture complementary cross-modal information. U2Fusion~\cite{xu2020u2fusion} adaptively preserves informative content from source images. SDNet~\cite{zhang2021sdnet} formulates fusion as gradient and intensity information extraction and reconstruction. CDDFuse~\cite{zhao2023cddfuse} decomposes features into base and detail components and models them with Transformer-based and INN-based branches, respectively. DeFusion~\cite{liang2022fusion} separates source images into common and unique feature embeddings for fusion. LRRNet~\cite{li2023lrrnet} introduces a lightweight fusion network guided by low-rank representation. SHIP~\cite{zheng2024probing} further models high-order synergistic interactions beyond pairwise cross-modal dependencies.

Another line exploits transformers and language guidance: Text-IF~\cite{yi2024text} introduces semantic text guidance for degradation-aware fusion, and FILM~\cite{zhao2024image} uses vision-language models to guide image fusion. MURF~\cite{xu2023murf}, TC-MoA~\cite{zhu2024task}, and S4Fusion~\cite{ma2025s4fusion} address misalignment, multi-task adaptation, and long-range dependency modeling, respectively.
% Another line of IVIF methods exploits transformers and language guidance to enhance cross-modal interaction. MURF~\cite{xu2023murf} jointly performs multi-modal registration and fusion, with the two tasks mutually optimized to cope with cross-modal misalignment. Text-IF~\cite{yi2024text} introduces semantic text guidance for degradation-aware and interactive image fusion. FILM~\cite{zhao2024image} leverages vision-language models to generate and fuse textual semantic descriptions, which then guide visual feature fusion via cross-attention. TC-MoA~\cite{zhu2024task} employs task-customized mixture-of-adapters to adapt a pretrained foundation model to multiple image fusion tasks. S4Fusion~\cite{ma2025s4fusion} further introduces selective state space modeling for efficient long-range cross-modal dependency modeling.

Beyond feed-forward reconstruction models, some methods exploit generative priors for IVIF. FusionGAN~\cite{ma2019fusiongan} uses adversarial learning to preserve visible textures and infrared targets. 
DCINN~\cite{wang2024general} adopts a conditional invertible network to perform detail-preserving fusion. 
DDFM~\cite{zhao2023ddfm} injects cross-modal information into a pretrained diffusion process, while Dif-Fusion~\cite{yue2023dif} extends diffusion-based fusion to multi-channel generation for better color fidelity. 
However, diffusion-based methods usually require higher inference costs.

\noindent\textbf{Fusion for downstream tasks.}

A parallel line of work couples image fusion with downstream perception tasks, making fused images more beneficial for high-level visual understanding. 
TarDAL~\cite{liu2022target} jointly formulates fusion and object detection via target-aware dual adversarial learning, and SegMIF~\cite{liu2023multi} builds an interactive fusion-segmentation framework. MetaFusion~\cite{zhao2023metafusion}, TIMFusion~\cite{liu2024task}, and DCEvo~\cite{liu2025dcevo} further incorporate meta-features, task guidance, and evolutionary learning into fusion.
% TarDAL~\cite{liu2022target} jointly formulates fusion and object detection through target-aware dual adversarial learning. SegMIF~\cite{liu2023multi} builds an interactive fusion-segmentation framework, where modality and semantic features mutually guide each other. MetaFusion~\cite{zhao2023metafusion} embeds detection-derived meta-features into the fusion process to enhance task-aware representations. TIMFusion~\cite{liu2024task} introduces task guidance, implicit architecture search, and meta initialization for flexible image fusion. DCEvo~\cite{liu2025dcevo} further uses evolutionary learning and cross-dimensional embedding to coordinate fusion quality and perception accuracy. 
These methods demonstrate the potential of task-aware optimization for bridging low-level fusion and high-level perception. Despite their varied designs, the above IVIF methods all build on the ANN paradigm, where cross-modal interaction is carried out through dense continuous activations.

\subsection{Deep Spiking Neural Networks}

\noindent\textbf{Deep SNN training and architecture.}

Spiking neural networks (SNNs) differ from artificial neural networks (ANNs) by communicating with discrete spikes. 
Each neuron accumulates input currents in its membrane potential and fires only when it exceeds a threshold~\cite{maass1997networks}. Spike-based communication provides sparsity and temporal dynamics, but the non-differentiable firing function makes deep SNN training challenging. To enable end-to-end optimization, surrogate gradient learning~\cite{neftci2019surrogate} approximates the gradient of spike activations and allows SNNs to be trained with spatio-temporal backpropagation~\cite{wu2019direct}. Based on this, directly trained deep SNNs have been advanced by temporal normalization~\cite{zheng2021going}, learnable neuron dynamics~\cite{fang2021incorporating}, and spiking residual architectures~\cite{fang2021deep}, which improve training stability and representation capacity.

\noindent\textbf{SNNs for visual representation and dense prediction.}

With advances in training and architecture design, SNNs have been extended from image classification to more complex visual tasks, including object detection~\cite{li2025brain}, optical flow estimation~\cite{hagenaars2021self}, and image reconstruction~\cite{zhu2022event}. This trend is particularly natural for neuromorphic cameras~\cite{xiao2025event,xiao2026learning, zhao2022learning, zhao2024boosting, wang2025sample, zhao2026spike}, which output sparse and asynchronous event or spike streams that align well with spike-driven computation~\cite{zheng2026spikecv,zhao2023spike,zhan2024spiking,zhang2025spiking}. SNNs provide temporal accumulation and sparse feature selection through membrane-potential dynamics~\cite{li2024spiking,li2025spiking}. 
However, their binary spike communication may suppress fine-grained responses that remain below the firing threshold, which can be particularly detrimental to IVIF because weak infrared targets and subtle visible textures may still provide valuable complementary evidence. The membrane potential retains such responses before spike firing, yet exploiting this continuous internal state for cross-modal fusion remains largely unexplored.
% This integrative nature is well-suited to aggregating complementary cues from the two modalities, yet leveraging it to jointly preserve infrared targets and visible textures in IVIF remains largely unexplored.

% SNNs provide temporal accumulation and sparse feature selection through membrane-potential dynamics. 
% However, how to leverage these dynamics to jointly preserve infrared targets and visible textures in IVIF remains underexplored.

\begin{figure}[t!]
    \centering
    \includegraphics[width=.9\linewidth]{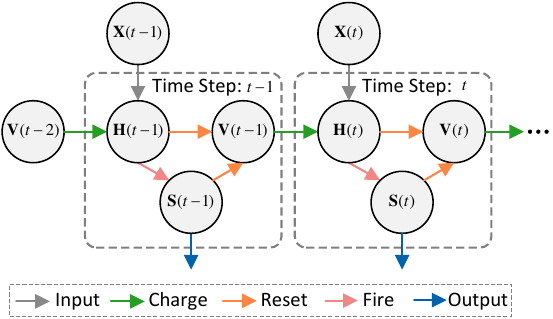}
    \vspace*{-6pt}
    \caption{Computational graph of a spiking neuron unrolled across two consecutive time steps~\cite{fang2021incorporating}. At each step $t$, the neuron receives an input $\mathbf{X}(t)$, charges its membrane potential from the previous state $\mathbf{V}(t{-}1)$ to an intermediate state $\mathbf{H}(t)$, fires a spike $\mathbf{S}(t)$ when the threshold is crossed, and then resets to obtain $\mathbf{V}(t)$ for the next step.} 
    \label{fig:snn_unroll}
    \vspace*{-6pt}
\end{figure}

\begin{figure}[t!]
    \centering
    \includegraphics[width=.96\linewidth]{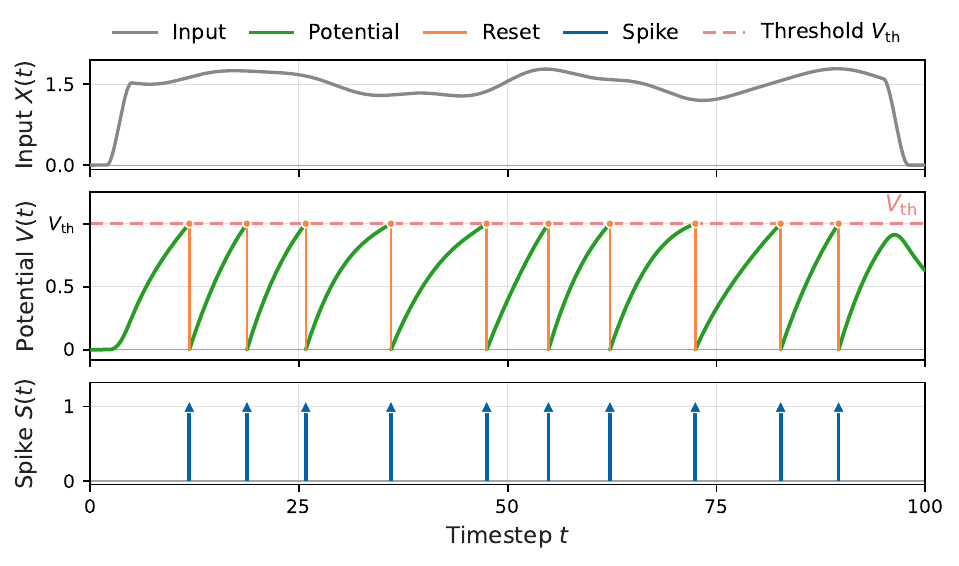}
    \vspace*{-10pt}
    \caption{Dynamics of a leaky integrate-and-fire (LIF) neuron under a constant input current. The input $X(t)$ continuously charges the membrane potential $V(t)$, which decays toward the resting potential in the absence of input. A spike $S(t)$ is emitted whenever $V(t)$ reaches the firing threshold $V_{\text{th}}$, after which the membrane potential is reset.}
    \label{fig:lif_dynamics}
    \vspace*{-10pt}
\end{figure}

\begin{figure*}[t!]
    \centering
    \includegraphics[width=.85\linewidth]{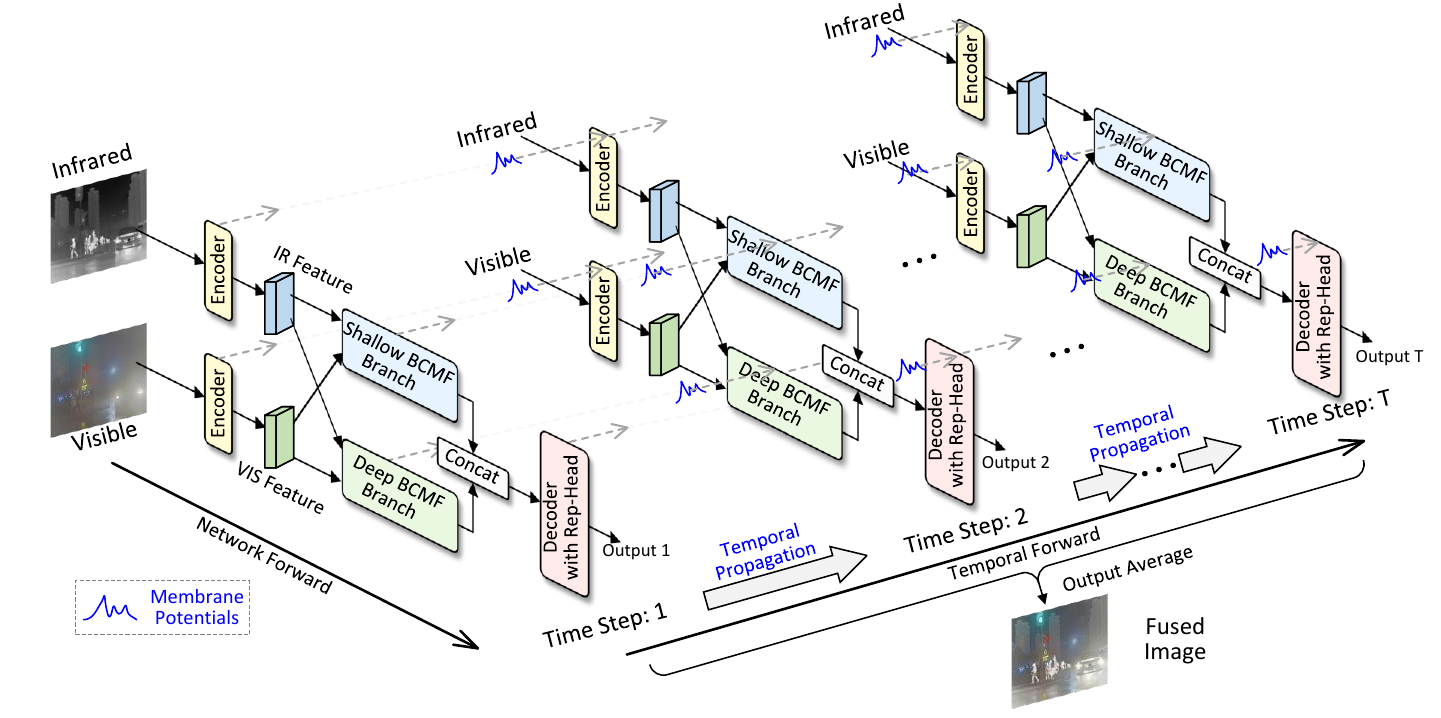}
    \vspace*{-6pt}
    \caption{Overall architecture of CIS-Fuse and its dual-axis forward propagation. Along the network forward axis, infrared and visible inputs are first encoded into IR and VIS features via a weight-shared encoder, then jointly processed by a shallow BCMF branch and a deep BCMF branch with asymmetric stacking depths. The two branch outputs are concatenated and decoded by a RepHead to produce the per-step output. Along the temporal propagation axis, the network unrolls over $T$ time steps, and the membrane potentials of all spiking neurons are propagated forward along the gray dashed arrows to carry temporal context. The outputs from all $T$ steps are averaged to yield the final fused image.}
    \label{fig:overall}
    \vspace*{-6pt}
\end{figure*}

\section{The proposed CIS-Fuse}

\subsection{Preliminaries of SNNs}
As the core of CIS-Fuse operates at the membrane potential, we first review the membrane-potential dynamics of a spiking neuron and fix the notation used throughout. Spiking neural networks process information through binary spikes over discrete time steps. 
For a spiking neuron at time step $t$, the membrane potential first charges from its previous state and the current input, then fires a binary spike once it crosses the threshold, and is finally reset according to the firing result. We describe these three steps as~\cite{wu2019direct,zhao2023spike}:
\begin{align}
\mathbf{H}(t) &= \mathcal{F}(\mathbf{V}(t-1), \mathbf{X}(t)), \label{eq:charge} \\
\mathbf{S}(t) &= \Theta(\mathbf{H}(t) - V_\text{th}), \label{eq:fire} \\
% \mathbf{V}(t) &= \mathbf{H}(t)(1 - \mathbf{S}(t)) + V_\text{reset}\mathbf{S}(t), \label{eq:reset}
\mathbf{V}(t) &= \mathbf{H}(t) - V_\text{th}\mathbf{S}(t), \label{eq:reset}
\end{align}
where $\mathbf{X}(t)$ is the input feature and $\mathbf{S}(t)$ is the output spike. 
$\mathbf{H}(t)$ denotes the pre-reset membrane potential, while $\mathbf{V}(t)$ is the post-reset state propagated to the next time step. 
The step function $\Theta(\cdot)$ outputs 1 when the membrane potential reaches the threshold $V_\text{th}$ and outputs 0 otherwise. 
Eq.~\eqref{eq:reset} adopts a soft reset that subtracts the threshold $V_\text{th}$ after firing, as opposed to a hard reset that clamps the potential to a fixed value $V_\text{reset}$. We use the soft reset throughout CIS-Fuse.
The function $\mathcal{F}(\cdot)$ specifies how temporal information is accumulated in the membrane potential, and different choices of $\mathcal{F}$ lead to different spiking neuron models. 
Fig.~\ref{fig:snn_unroll} illustrates the recurrent computation of this process, and Fig.~\ref{fig:lif_dynamics} shows the membrane evolution of a LIF neuron under a constant input current.

The integrate-and-fire (IF) neuron adopts the simplest charging function, in which the membrane potential perfectly accumulates all past inputs:
\begin{equation}
\mathcal{F}_\text{IF}(\mathbf{V}(t-1), \mathbf{X}(t)) = \mathbf{V}(t-1) + \mathbf{X}(t).
\label{eq:if}
\end{equation}
While computationally simple, the IF neuron lacks any forgetting mechanism, so past inputs accumulate indefinitely without decay. The leaky integrate-and-fire (LIF) neuron addresses this by introducing an exponential leakage term controlled by the membrane time constant $\tau$:
\begin{equation}
\begin{aligned}
&\mathcal{F}_\text{LIF}(\mathbf{V}(t-1), \mathbf{X}(t)) \\
&= \mathbf{V}(t-1) + \frac{1}{\tau}\bigl(-(\mathbf{V}(t-1) - V_\text{reset}) + \mathbf{X}(t)\bigr),
\end{aligned}
\label{eq:lif}
\end{equation}
where a larger $\tau$ produces longer memory while a smaller $\tau$ emphasizes recent inputs. In standard LIF, $\tau$ is a fixed hyperparameter shared across all neurons, which limits the model's ability to adapt temporal dynamics to data. The parametric LIF (PLIF) neuron~\cite{fang2021incorporating} addresses this limitation by parameterizing the leakage as $1/\tau = \sigma(w)$ with $w$ a learnable scalar and $\sigma(\cdot)$ the sigmoid function, so that $1/\tau$ remains in $(0, 1)$ throughout training. Each spiking layer thus learns its own temporal time scale jointly with the network weights, which has been shown to substantially improve representation capacity in deep SNNs. We adopt PLIF neurons throughout CIS-Fuse for this reason. For notational simplicity, we omit the time-step superscript $(t)$ from feature, current, membrane-potential, and spike variables in the subsequent network-level formulations, unless the temporal index is explicitly required.

\subsection{Overall Architecture of the CIS-Fuse}
\label{sec:overall}
The overall architecture of CIS-Fuse is illustrated in Fig.~\ref{fig:overall}, which unrolls along two orthogonal axes: the network forward axis that processes spatial features from the two modalities, and the temporal propagation axis that evolves the membrane potentials across $T$ time steps.

\begin{figure}[t!]
    \centering
    \includegraphics[width=.8\linewidth]{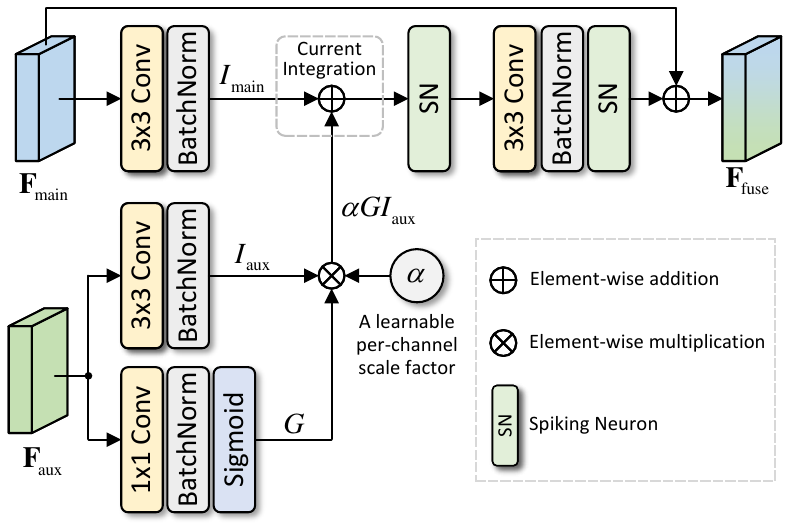}
    \vspace*{-4pt}
    \caption{Structure of the proposed Current Injection Spiking (CIS) module. The main and auxiliary features are encoded into two currents, with the auxiliary one additionally modulated by a content-dependent gate and a learnable per-channel scale factor. 
    The two currents are combined before thresholding to jointly charge the same membrane potential.
    % The two currents are integrated at the membrane potential level before spike firing.
    }
    \vspace*{-10pt}
    \label{fig:cis_module}
\end{figure}

\textbf{Network forward.} Along the forward axis, the infrared image $\mathbf{I}_\text{ir}$ and the visible image $\mathbf{I}_\text{vis}$ are first fed into a weight-shared spiking encoder, which transforms each modality into a spike-based feature. We denote the resulting features as the IR feature $\mathbf{F}_\text{ir}$ and the VIS feature $\mathbf{F}_\text{vis}$. The two features are then jointly processed by two parallel fusion branches with asymmetric stacking depths: a shallow BCMF branch with a single BCMF block and a deep BCMF branch stacking three such blocks of the same structure but with independent parameters. Each BCMF block, built upon the proposed CIS operator, performs bidirectional cross-modal current injection between IR and VIS, and each branch outputs a fused spiking feature. The two branch outputs are concatenated along the channel dimension and passed to a spiking decoder equipped with a reparameterized head (abbreviated as RepHead hereafter), which produces the per-step output.

\textbf{Temporal propagation.} Since CIS-Fuse is a spiking neural network, all neurons maintain a membrane potential that carries temporal context across time steps. The network is unrolled for $T$ steps with shared parameters: at every step, the same pair of input images is fed in, while the membrane potentials of all spiking neurons are inherited from the previous step. For static inputs, the temporal axis represents iterative membrane charging over repeated presentations rather than physical scene evolution. This allows the network to progressively refine its internal representations. The $T$ per-step outputs are aggregated by averaging to produce the final fused image. 
% Since the source images are static, the temporal axis here does not represent physical scene evolution. Instead, repeated presentation over $T$ steps provides an iterative charging process in which membrane potentials accumulate and refine the evidence extracted from the same input pair.
% :
% \begin{equation}
% \label{eq:Ifused}
% \vspace*{-4pt}
% \mathbf{I}_\text{fused} = \frac{1}{T}\sum_{t=1}^{T} \mathcal{N}(\mathbf{I}_\text{ir}, \mathbf{I}_\text{vis}; t),
% \label{eq:output_avg}
% \vspace*{-4pt}
% \end{equation}
% where $\mathcal{N}(\cdot;t)$ denotes the network output at time step $t$. 

% This output-mean supervision ties the training loss to the temporally aggregated prediction rather than any single step, encouraging the network to leverage its full $T$-step dynamics during training.

In the following subsections, we detail the CIS operator and its bidirectional extension into the BCMF module and the dual-branch deployment in Sec.~\ref{sec:cis_module}. The RepHead reparameterization strategy is in Sec.~\ref{sec:rephead}.

\subsection{Current Injection Module}
\label{sec:cis_module}
The two modalities are encoded into features $\mathbf{F}_\text{ir}$ and $\mathbf{F}_\text{vis}$ by the weight-shared encoder, whose body is a patch-embedding convolution followed by a stack of ResBlockSNN blocks. We build the proposed cross-modal fusion mechanism to process these features, organized in two levels: the bottom-level \textbf{CIS module} performs unidirectional cross-modal current injection, while the upper-level \textbf{BCMF block} pairs two CIS modules into bidirectional modulation with alternating refinement.

\textbf{CIS module.} The CIS module takes a pair of features $(\mathbf{F}_\text{main}, \mathbf{F}_\text{aux})$ as input, where $\mathbf{F}_\text{main}$ serves as the main driving modality and $\mathbf{F}_\text{aux}$ as the auxiliary modulating modality. The two features are first encoded into input currents $\mathbf{I}_\text{main}$ and $\mathbf{I}_\text{aux}$ through a $3\times3$ Conv-BN block, respectively. A content-dependent gate $\mathbf{G}$ is computed from $\mathbf{F}_\text{aux}$ itself via a $1\times1$ Conv-BN-Sigmoid branch. 
% The main current and the gated auxiliary current are then integrated at the membrane potential level to form the input current of the spiking neuron:
\rvs{The main current and the gated auxiliary current are combined before spike generation to form a fused current that charges the membrane potential:
\begin{equation}
\mathbf{I}_{\text{cis}}
=
\mathbf{I}_{\text{main}}
+
\boldsymbol{\alpha}
\mathbf{G}
\mathbf{I}_{\text{aux}},
\label{eq:cis_integration}
\end{equation}
% where $\boldsymbol{\alpha} \in \mathbb{R}^{C \times 1 \times 1}$ is a learnable per-channel injection scale factor, and the multiplication among $\boldsymbol{\alpha}$, $\mathbf{G}$, and $\mathbf{I}_\text{aux}$ is performed element-wise. The integrated current is first passed through a spiking neuron and then a $3\times3$ Conv-BN-SN block, with a residual connection from $\mathbf{F}_\text{main}$ giving the fused output:
where $\boldsymbol{\alpha}\in\mathbb{R}^{C\times1\times1}$ is a learnable per-channel injection scale factor, and the multiplication among $\boldsymbol{\alpha}$, $\mathbf{G}$, and $\mathbf{I}_{\text{aux}}$ is performed element-wise. This fused current $\mathbf{I}_{\text{cis}}$ plays the role of the input $\mathbf{X}(t)$ in Eq.~\eqref{eq:charge}.

Following the charging and firing dynamics of a spiking neuron, the fused current updates the pre-reset membrane potential and determines the emitted spike as
\begin{align}
\mathbf{H}_{\text{fuse}}
&=
\mathcal{F}
\left(
\mathbf{V}_{\text{fuse}},
\mathbf{I}_{\text{cis}}
\right),
\label{eq:cis_membrane_h}
\\
\mathbf{S}_{\text{fuse}}
&=
\Theta
\left(
\mathbf{H}_{\text{fuse}}
-
V_{\text{th}}
\right),
\label{eq:cis_membrane_s}
\end{align}
where $\mathbf{V}_{\text{fuse}}$ denotes the membrane state inherited from the preceding time step. Therefore, an auxiliary response that is insufficient to trigger a spike independently can still shift the membrane potential and influence the firing decision when combined with the main current. In this sense, cross-modal interaction occurs through the continuous membrane potential before being converted into a binary spike.}

The resulting spike is subsequently processed by a $3\times3$ Conv-BN-SN block, with a residual connection from $\mathbf{F}_{\text{main}}$ giving the fused output:
\begin{equation}
\mathbf{F}_{\text{fuse}}
=
\text{SN}
\left(
\text{BN}
\left(
\text{Conv}_{3\times3}
\left(
\mathbf{S}_{\text{fuse}}
\right)
\right)
\right)
+
\mathbf{F}_{\text{main}}.
\label{eq:cis_output}
\end{equation}

% \begin{equation}
% \mathbf{F}_\text{fuse} = \text{SN}(\text{BN}(\text{Conv}_{3\times3}(\text{SN}(\mathbf{I}_\text{fuse})))) + \mathbf{F}_\text{main}.
% \label{eq:cis_output}
% \end{equation}

\begin{figure}[t!]
    \centering
    \includegraphics[width=.7\linewidth]{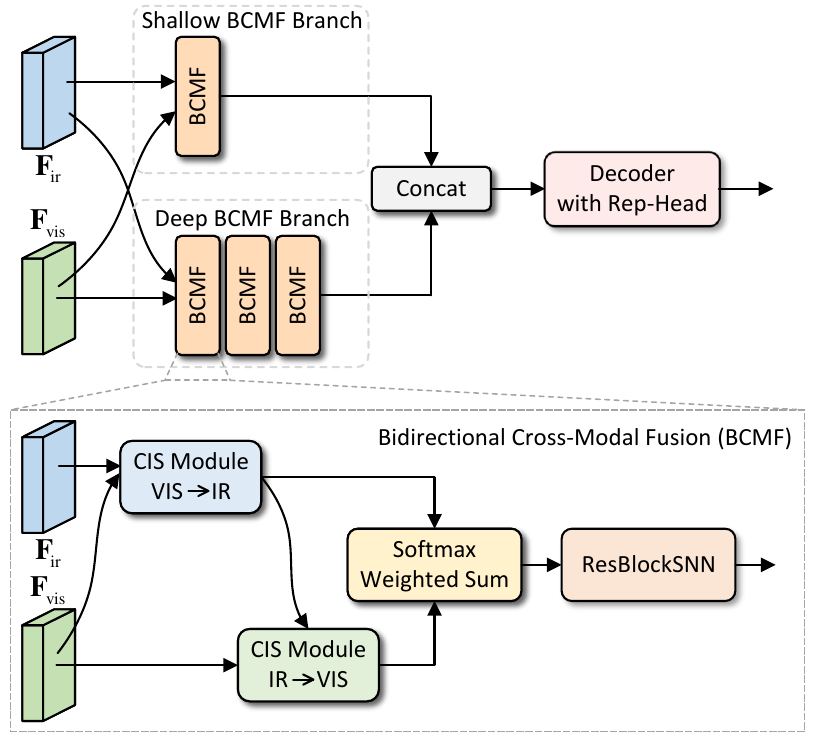}
    \vspace*{-4pt}
    \caption{Structure of a Bidirectional Cross-Modal Fusion (BCMF) branch. The branch stacks $K$ blocks, each carrying out one round of alternating bidirectional injection between IR and VIS through a pair of CIS modules. Only the last block is followed by a softmax-weighted sum and a refinement block, while the preceding blocks contain only the alternating CIS pair. The shallow and deep BCMF branches use $K{=}1$ and $K{=}3$, respectively.}
    \vspace*{-4pt}
    \label{fig:bcmf}
\end{figure}
\begin{figure}
    \centering
    \includegraphics[width=.6\linewidth]{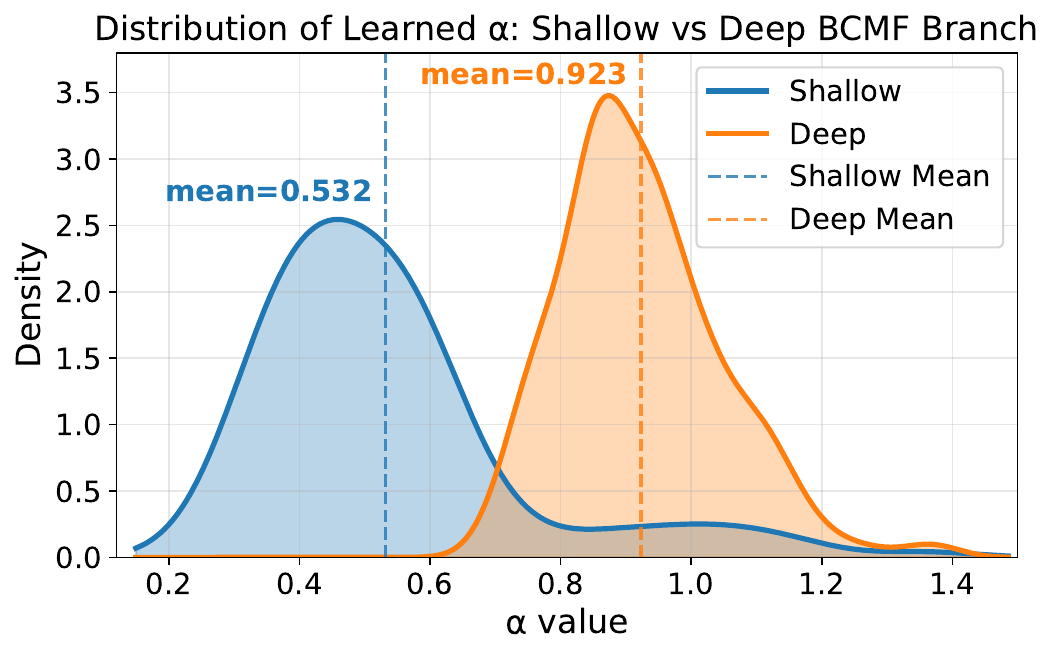}
    \vspace*{-10pt}
    \caption{Distribution of the learned per-channel injection strength $\alpha$ in the shallow and deep BCMF branches after training. The two distributions are substantially separated, indicating that, when left to optimize freely, the network settles into a clear functional specialization between the two branches.}
    \vspace*{-12pt}
    \label{fig:alpha_dist}
\end{figure}

For brevity, we denote the entire CIS module as
$\mathbf{F}_{\text{fuse}}
=
\text{CIS}
(\mathbf{F}_{\text{main}},\mathbf{F}_{\text{aux}})$.
Two design components distinguish CIS from generic cross-attention or gated fusion: (i) a driver-modulator asymmetry in which only the auxiliary current is gated, preserving the integrity of the main modality; and (ii) the per-channel learnable $\boldsymbol{\alpha}$ that allows different feature channels to adaptively regulate the modulation magnitude. More importantly, the two modality-specific currents are combined before thresholding and jointly contribute to the membrane-potential dynamics in Eq.~\eqref{eq:cis_membrane_h} and \eqref{eq:cis_membrane_s}. This mechanism also draws inspiration from dendritic integration in biological neurons~\cite{london2005dendritic}, where currents arriving from multiple sources are integrated before the soma generates a spike.
% For brevity, we denote the entire CIS module as $\mathbf{F}_\text{fuse} = \text{CIS}(\mathbf{F}_\text{main}, \mathbf{F}_\text{aux})$. Two design components distinguish CIS from a generic cross-attention or gated fusion: (i) a driver-modulator asymmetry in which only the auxiliary current is gated, preserving the integrity of the main modality; and (ii) the per-channel learnable $\boldsymbol{\alpha}$ that allows different feature channels to adaptively regulate the modulation magnitude. The additive integration in Eq.~\eqref{eq:cis_integration}, together with the asymmetric treatment of the two currents, echoes in spirit the dendritic integration mechanism of biological neurons~\cite{london2005dendritic}, where multiple input currents from different sources are summed at the membrane potential before the soma decides whether to fire.

\textbf{BCMF block.} A BCMF block builds on CIS to realize bidirectional cross-modal modulation with alternating refinement, and a branch is formed by stacking $K$ such blocks. With $(\mathbf{F}_\text{ir}^{(0)}, \mathbf{F}_\text{vis}^{(0)}) = (\mathbf{F}_\text{ir}, \mathbf{F}_\text{vis})$, the $k$-th block updates the two modalities by:
\begin{align}
\mathbf{F}_\text{ir}^{(k)} &= \text{CIS}_a^{(k)}\bigl(\mathbf{F}_\text{ir}^{(k-1)},\, \mathbf{F}_\text{vis}^{(k-1)}\bigr), \label{eq:bcmf_ir} \\
\mathbf{F}_\text{vis}^{(k)} &= \text{CIS}_b^{(k)}\bigl(\mathbf{F}_\text{vis}^{(k-1)},\, \mathbf{F}_\text{ir}^{(k)}\bigr). \label{eq:bcmf_vis}
\end{align}
Note that the second equation conditions on the \emph{just-updated} $\mathbf{F}_\text{ir}^{(k)}$ rather than $\mathbf{F}_\text{ir}^{(k-1)}$, which is the key mechanism of alternating refinement: each VIS update is built upon an IR feature that has already absorbed the current round of VIS information. The two CIS modules inside a block, as well as those across different blocks, have independent parameters, so the $K$ blocks form a stacked sequence rather than a weight-shared recurrence. After the $K$ blocks, the two refined features $\mathbf{F}_\text{ir}^{(K)}$ and $\mathbf{F}_\text{vis}^{(K)}$ are merged through a softmax-weighted sum and passed to a spiking residual block:
\begin{equation}
\mathbf{F}_\text{bcmf} = \text{ResBlockSNN}\bigl(\mathbf{W}_1 \mathbf{F}_\text{ir}^{(K)} + \mathbf{W}_2 \mathbf{F}_\text{vis}^{(K)}\bigr),
\label{eq:bcmf_output}
\end{equation}
where $\mathbf{W}_1$ and $\mathbf{W}_2$ are pixel-wise softmax weights predicted from $[\mathbf{F}_\text{ir}^{(K)}, \mathbf{F}_\text{vis}^{(K)}]$ by a lightweight Conv-BN-ReLU-Conv head, and the multiplication between weights and features is element-wise. We denote the resulting branch as $\mathbf{F}_\text{bcmf} = \text{BCMF}^K(\mathbf{F}_\text{ir}, \mathbf{F}_\text{vis})$, where the superscript $K$ is the number of stacked blocks.

\textbf{Dual-branch deployment.} We deploy two parallel branches on the same encoded pair $(\mathbf{F}_\text{ir}, \mathbf{F}_\text{vis})$: a shallow BCMF branch $\text{BCMF}^{1}$ with a single block and a deep BCMF branch $\text{BCMF}^{3}$ with three stacked blocks. Since the blocks carry independent parameters, the deep BCMF branch is three blocks in sequence rather than one block reused three times. The two branches are initialized with a soft asymmetric bias on the injection strength $\alpha$ but are otherwise free to be updated by the training objective, with no hard architectural constraints on their parameters. After training, the learned $\alpha$ distributions of the two branches become substantially separated, as visualized in Fig.~\ref{fig:alpha_dist}. The shallow BCMF branch settles into a regime of mild cross-modal modulation, while the deep BCMF branch performs much stronger modulation. This indicates that, when left to optimize freely, the network settles into 
% a clear functional specialization between the two branches, an emergent division of labor that echoes the magnocellular and parvocellular pathway organization of the biological visual system~\cite{livingstone1988segregation}, where parallel pathways with distinct response characteristics jointly process complementary aspects of visual information. 
a clear functional specialization between the two branches, echoing the parallel magnocellular and parvocellular pathways of the biological visual system~\cite{livingstone1988segregation}.
The two branch outputs are concatenated along the channel dimension and forwarded to the decoder.

\begin{figure}
    \centering
    \includegraphics[width=.75\linewidth]{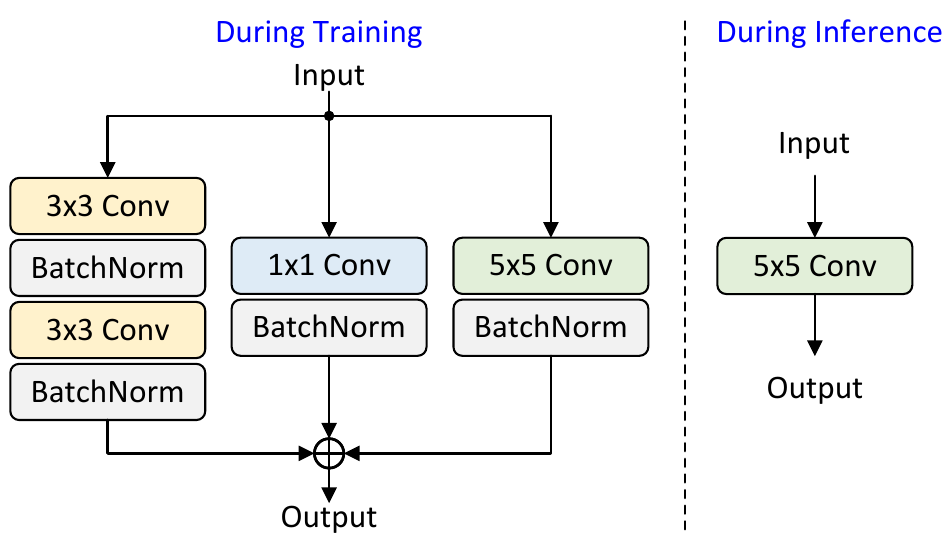}
    \caption{Structure of the proposed RepHead. \textit{Left}: the training-time multi-branch design with parallel 3$\times$3, 1$\times$1, and 5$\times$5 Conv-BN branches summed element-wise. \textit{Right}: the inference-time equivalent, obtained by algebraically collapsing the three branches into a single 5$\times$5 convolution.}
    \label{fig:rephead}
    \vspace*{-12pt}
\end{figure}

\subsection{Reparameterized Output Head} \label{sec:rephead} At each time step, the decoder maps the fused spiking feature to a fused image. Since this final projection bridges spike-based representations and pixel intensities, a single convolution may limit reconstruction quality, whereas directly deepening the head increases inference cost. To resolve this tension, the RepHead adopts a structurally reparameterized design~\cite{ding2021repvgg, zhao2023spike}, shown in Fig.~\ref{fig:rephead}. The key idea is to decouple training-time expressiveness from inference-time cost: RepHead uses a multi-branch design during training, and is algebraically collapsed into a single convolution at inference. We next describe its two forms in turn. 

\textbf{Training-time multi-branch structure.} Given the decoder feature $\mathbf{F}$ at each time step, RepHead processes it with three parallel Conv-BN branches: a two-layer $3{\times}3$ branch, a single $1{\times}1$ branch, and a single $5{\times}5$ branch. These branches provide complementary parameterizations, where the stacked $3{\times}3$ convolutions model local spatial compositions, the $1{\times}1$ branch preserves channel-wise mixing, and the $5{\times}5$ branch introduces a larger spatial receptive field. Since no nonlinear activation is inserted inside the branches, each branch remains a linear transformation of the input feature. The training-time output is obtained by element-wise summation: \begin{equation} \mathcal{H}_{\mathrm{train}}(\mathbf{F}) = \mathcal{B}_{3\times 3}(\mathbf{F}) + \mathcal{B}_{1\times 1}(\mathbf{F}) + \mathcal{B}_{5\times 5}(\mathbf{F}), \end{equation} where $\mathcal{B}_{3\times 3}$, $\mathcal{B}_{1\times 1}$, and $\mathcal{B}_{5\times 5}$ denote the three training-time branches. This design enriches the optimization-time representation while preserving the linearity required for exact structural reparameterization.

\textbf{Inference-time fused convolution.} Since each branch is linear in $\mathbf{F}$ and BN can be folded into its preceding convolution, the entire RepHead is mathematically equivalent to a single convolution. We pre-compute the equivalent kernel by folding each Conv-BN pair, collapsing the two-layer $3{\times}3$ branch into an equivalent $5{\times}5$ kernel via convolution, zero-padding the $1{\times}1$ kernel to $5{\times}5$, and summing the three resulting kernels:
\begin{equation}
    \mathbf{K}_{\text{infer}} 
    = \mathbf{K}_{3\times 3\rightarrow 5\times 5}^{\text{fold}} 
    + \text{pad}_{5}\!\left(\mathbf{K}_{1\times 1}^{\text{fold}}\right)
    + \mathbf{K}_{5\times 5}^{\text{fold}}.
\end{equation}
At inference, the entire RepHead therefore reduces to a single $5{\times}5$ convolution. From a deployment perspective, this means the deployed RepHead is structurally identical to a plain convolutional layer, and thus imposes no additional burden on SNN inference pipelines.

% ===== Ranking highlight macros =====
% 1st/2nd: green; 3rd/4th: yellow.
\definecolor{rankonebg}{rgb}{.776,.937,.808}
\definecolor{rankonefg}{rgb}{0,.38,0}

\definecolor{ranktwobg}{rgb}{.886,.937,.855}
\definecolor{ranktwofg}{rgb}{.216,.337,.137}

\definecolor{rankthreebg}{rgb}{1,.949,.8}
\definecolor{rankthreefg}{rgb}{.502,.376,0}

\definecolor{rankfourbg}{rgb}{1,.902,.6}
\definecolor{rankfourfg}{rgb}{.498,.376,0}

\newcommand{\RankCell}[3]{%
  \cellcolor{#1}\textcolor{#2}{\bfseries #3}%
}

\newcommand{\RankOne}[1]{\RankCell{rankonebg}{rankonefg}{#1}}
\newcommand{\RankTwo}[1]{\RankCell{ranktwobg}{ranktwofg}{#1}}
\newcommand{\RankThree}[1]{\RankCell{rankthreebg}{rankthreefg}{#1}}
\newcommand{\RankFour}[1]{\RankCell{rankfourbg}{rankfourfg}{#1}}
\newcommand{\venuesize}[0]{\tiny}

\renewcommand{\arraystretch}{1.0}
\setlength\tabcolsep{4pt}
\begin{table*}[htbp]
  \centering
  \footnotesize

    \caption{Quantitative comparison on RoadScene, MSRS, M3FD, and FMB benchmarks. 
    For each metric, the top four results are marked from green to yellow, as detailed at the bottom. AvgRank denotes the average ranking across all evaluation metrics.}
    \vspace{-4pt}
    \label{tab:quantitative_comparison}
    
    \begin{tabular}{@{}l@{}}
    
    \begin{tabular}{lcccccccc||cccccccc}
        \toprule
        \multicolumn{1}{c}{Benchmark} & \multicolumn{8}{c||}{RoadScene} & \multicolumn{8}{c}{MSRS} \\
        \midrule
        \multicolumn{1}{c}{Method / Metric} & EN & SD & SF & AG & SCD & CC & VIFF & AvgRank & EN & SD & SF & AG & SCD & CC & VIFF & AvgRank \\
        \midrule
        SDNet~\cite{zhang2021sdnet}~{\venuesize \textcolor{venuedark}{[IJCV'21]}} & 7.14 & 40.20 & 13.70 & 5.36 & 1.49 & 0.55 & 0.60 & {9.4} & 5.25 & 17.35 & 8.67 & 2.67 & 0.99 & \RankTwo{0.64} & 0.50 & 12.6 \\
        U2Fusion~\cite{xu2020u2fusion}~{\venuesize \textcolor{venuedark}{[TPAMI'22]}} & 7.09 & 38.12 & 13.25 & 5.27 & \RankTwo{1.70} & \RankOne{0.61} & 0.60 & {8.6} & 5.37 & 25.52 & 9.07 & 2.82 & 1.24 & \RankOne{0.65} & 0.54 & 10.9 \\
        TarDAL~\cite{liu2022target}~{\venuesize \textcolor{venuedark}{[CVPR'22]}} & 7.17 & \RankThree{47.44} & 10.83 & 3.94 & 1.55 & 0.56 & 0.54 & {9.7} & 5.28 & 25.22 & 5.98 & 1.83 & 0.71 & 0.45 & 0.42 & 15.4 \\
        DeFusion~\cite{liang2022fusion}~{\venuesize \textcolor{venuedark}{[ECCV'22]}} & \RankFour{7.23} & 44.44 & 10.22 & 3.99 & \RankThree{1.69} & 0.58 & 0.63 & {8.6} & 6.46 & 37.63 & 8.60 & 2.80 & 1.35 & 0.60 & 0.77 & 10.4 \\
        LRRNet~\cite{li2023lrrnet}~{\venuesize \textcolor{venuedark}{[TPAMI'23]}} & 6.99 & 40.00 & 11.14 & 4.02 & 1.61 & 0.58 & 0.49 & {11.4} & 6.19 & 31.78 & 8.46 & 2.63 & 0.79 & 0.51 & 0.54 & 13.4 \\
        MURF~\cite{xu2023murf}~{\venuesize \textcolor{venuedark}{[TPAMI'23]}} & 6.83 & 36.73 & \RankOne{16.20} & \RankOne{6.18} & 1.56 & \RankTwo{0.59} & 0.54 & {8.3} & 5.04 & 20.63 & 10.49 & 3.38 & 1.02 & 0.62 & 0.44 & 11.9 \\
        Dif-Fusion~\cite{yue2023dif}~{\venuesize \textcolor{venuedark}{[TIP'23]}} & 7.16 & 43.71 & 14.12 & \RankThree{5.56} & 1.43 & 0.54 & 0.59 & {8.6} & \RankThree{6.66} & 41.90 & \RankThree{11.63} & \RankTwo{3.88} & 1.59 & 0.60 & 0.83 & \RankFour{4.7} \\
        SegMIF~\cite{liu2023multi}~{\venuesize \textcolor{venuedark}{[ICCV'23]}} & 7.16 & 41.88 & 13.52 & 5.08 & 1.60 & \RankThree{0.59} & 0.67 & \RankFour{7.1} & 5.95 & 37.28 & 11.10 & 3.47 & 1.57 & \RankThree{0.63} & 0.88 & 7.4 \\
        TIMFusion~\cite{liu2024task}~{\venuesize \textcolor{venuedark}{[TPAMI'24]}} & 7.18 & \RankFour{46.56} & 12.87 & 4.71 & 1.30 & 0.53 & \RankThree{0.74} & {9.3} & 6.27 & 36.73 & 9.67 & 2.91 & 1.34 & 0.58 & 0.65 & 11.1 \\
        DCINN~\cite{wang2024general}~{\venuesize \textcolor{venuedark}{[IJCV'24]}} & 7.00 & 38.24 & 10.28 & 4.06 & 1.32 & 0.53 & 0.23 & {14.3} & 6.00 & 40.30 & 10.51 & 3.34 & 1.49 & 0.58 & 0.82 & 9.7 \\
        SHIP~\cite{zheng2024probing}~{\venuesize \textcolor{venuedark}{[CVPR'24]}} & 7.15 & 44.67 & \RankThree{14.58} & \RankFour{5.45} & 1.41 & 0.53 & 0.71 & {7.7} & 6.43 & 41.13 & \RankOne{11.81} & \RankOne{3.92} & 1.51 & 0.59 & \RankFour{0.91} & 5.7 \\
        TC-MoA~\cite{zhu2024task}~{\venuesize \textcolor{venuedark}{[CVPR'24]}} & 7.12 & 41.97 & 10.24 & 4.06 & 1.42 & 0.55 & 0.67 & {11.1} & 6.55 & 41.63 & 10.91 & 3.58 & \RankFour{1.60} & 0.61 & 0.81 & 6.4 \\
        Text-IF~\cite{yi2024text}~{\venuesize \textcolor{venuedark}{[CVPR'24]}} & \RankTwo{7.28} & \RankTwo{48.62} & 13.55 & 5.30 & \RankFour{1.67} & 0.58 & \RankFour{0.74} & \RankTwo{4.7} & \RankTwo{6.67} & \RankFour{42.60} & 11.46 & 3.70 & \RankThree{1.62} & 0.60 & \RankOne{1.04} & \RankThree{4.3} \\
        S4Fusion~\cite{ma2025s4fusion}~{\venuesize \textcolor{venuedark}{[TIP'25]}} & \RankThree{7.28} & 46.54 & \RankFour{14.54} & 5.41 & 1.36 & 0.52 & \RankOne{0.80} & \RankThree{6.9} & 6.53 & \RankThree{42.75} & \RankTwo{11.71} & \RankThree{3.79} & 1.53 & 0.59 & \RankThree{1.00} & 5.1 \\
        DCEvo~\cite{liu2025dcevo}~{\venuesize \textcolor{venuedark}{[CVPR'25]}} & 7.15 & 45.22 & 13.42 & 4.91 & 1.57 & 0.56 & \RankTwo{0.77} & {7.6} & \RankFour{6.63} & \RankTwo{42.94} & \RankFour{11.46} & 3.71 & \RankTwo{1.66} & 0.61 & \RankTwo{1.03} & \RankTwo{3.7} \\
        CIS-Fuse (Ours)~{\venuesize \textcolor{venuedark}{[Ours]}} & \RankOne{7.38} & \RankOne{53.32} & \RankTwo{15.28} & \RankTwo{5.89} & \RankOne{1.79} & \RankFour{0.59} & 0.66 & \RankOne{2.7} & \RankOne{6.69} & \RankOne{43.70} & 11.30 & \RankFour{3.74} & \RankOne{1.86} & \RankFour{0.62} & 0.90 & \RankOne{3.1} \\
        
        \midrule
        \multicolumn{1}{c}{Benchmark} & \multicolumn{8}{c||}{M3FD} & \multicolumn{8}{c}{FMB} \\
        \midrule
        \multicolumn{1}{c}{Method / Metric} & EN & SD & SF & AG & SCD & CC & VIFF & AvgRank & EN & SD & SF & AG & SCD & CC & VIFF & AvgRank \\
        \midrule
        SDNet~\cite{zhang2021sdnet}~{\venuesize \textcolor{venuedark}{[IJCV'21]}} & 6.84 & 35.10 & 13.60 & 4.72 & \RankFour{1.52} & 0.50 & 0.58 & {8.6} & 6.15 & 21.82 & 10.86 & 3.53 & 1.17 & 0.62 & 0.64 & 12.3 \\
        U2Fusion~\cite{xu2020u2fusion}~{\venuesize \textcolor{venuedark}{[TPAMI'22]}} & \RankTwo{6.96} & 35.29 & 13.12 & 4.85 & \RankOne{1.75} & \RankTwo{0.57} & 0.68 & \RankFour{5.6} & 6.60 & 29.15 & 10.31 & 3.41 & \RankFour{1.49} & \RankTwo{0.65} & 0.66 & 9.1 \\
        TarDAL~\cite{liu2022target}~{\venuesize \textcolor{venuedark}{[CVPR'22]}} & 6.87 & \RankOne{42.18} & 7.63 & 2.67 & 1.29 & 0.46 & 0.55 & {11.9} & 6.63 & \RankOne{39.34} & 6.94 & 2.17 & 1.03 & 0.51 & 0.56 & 12.7 \\
        DeFusion~\cite{liang2022fusion}~{\venuesize \textcolor{venuedark}{[ECCV'22]}} & 6.77 & 33.32 & 9.04 & 3.17 & 1.48 & 0.51 & 0.59 & {11.3} & \RankFour{6.71} & \RankFour{35.13} & 8.84 & 2.77 & 1.49 & 0.62 & 0.64 & 8.6 \\
        LRRNet~\cite{li2023lrrnet}~{\venuesize \textcolor{venuedark}{[TPAMI'23]}} & 6.44 & 27.17 & 10.68 & 3.59 & 1.46 & \RankFour{0.54} & 0.57 & {12.4} & 6.28 & 25.89 & 10.12 & 3.05 & 1.34 & \RankThree{0.64} & 0.61 & 12.4 \\
        MURF~\cite{xu2023murf}~{\venuesize \textcolor{venuedark}{[TPAMI'23]}} & 6.50 & 31.48 & 12.55 & 4.87 & 1.46 & 0.52 & 0.42 & {11.1} & 6.37 & 31.28 & 13.88 & \RankOne{4.77} & 1.34 & 0.60 & 0.45 & 9.9 \\
        Dif-Fusion~\cite{yue2023dif}~{\venuesize \textcolor{venuedark}{[TIP'23]}} & 6.80 & 33.86 & \RankFour{14.73} & \RankTwo{5.10} & 1.38 & 0.48 & 0.59 & {9.3} & 6.66 & 33.12 & 14.12 & 4.38 & 1.42 & 0.60 & 0.63 & 8.0 \\
        SegMIF~\cite{liu2023multi}~{\venuesize \textcolor{venuedark}{[ICCV'23]}} & 6.85 & 36.17 & 14.14 & 4.81 & \RankTwo{1.72} & \RankOne{0.57} & 0.74 & \RankThree{4.9} & \RankTwo{6.83} & \RankThree{37.08} & 13.69 & 4.18 & \RankOne{1.72} & \RankOne{0.66} & 0.78 & \RankTwo{4.0} \\
        TIMFusion~\cite{liu2024task}~{\venuesize \textcolor{venuedark}{[TPAMI'24]}} & 6.75 & 33.12 & 12.31 & 4.11 & 1.37 & 0.49 & 0.70 & {11.9} & 6.51 & 28.44 & 12.23 & 3.59 & 1.24 & 0.58 & 0.71 & 11.6 \\
        DCINN~\cite{wang2024general}~{\venuesize \textcolor{venuedark}{[IJCV'24]}} & 6.59 & 29.43 & 11.21 & 3.84 & 1.46 & 0.52 & 0.69 & {11.4} & 6.47 & 28.65 & 11.47 & 3.51 & 1.39 & 0.61 & 0.76 & 10.4 \\
        SHIP~\cite{zheng2024probing}~{\venuesize \textcolor{venuedark}{[CVPR'24]}} & 6.82 & 35.22 & \RankOne{15.26} & \RankOne{5.16} & 1.31 & 0.47 & \RankThree{0.82} & {7.3} & 6.66 & 34.09 & \RankThree{14.58} & \RankThree{4.44} & 1.35 & 0.58 & \RankThree{0.89} & 6.9 \\
        TC-MoA~\cite{zhu2024task}~{\venuesize \textcolor{venuedark}{[CVPR'24]}} & 6.84 & 34.22 & 13.57 & 4.50 & 1.47 & 0.50 & 0.77 & {8.6} & 6.66 & 33.07 & 13.44 & 3.97 & 1.43 & 0.59 & 0.81 & 8.3 \\
        Text-IF~\cite{yi2024text}~{\venuesize \textcolor{venuedark}{[CVPR'24]}} & \RankThree{6.90} & \RankFour{36.75} & \RankTwo{15.09} & \RankFour{5.08} & 1.49 & 0.50 & \RankTwo{0.90} & \RankTwo{4.6} & \RankThree{6.72} & 34.75 & \RankTwo{14.66} & \RankFour{4.39} & 1.48 & 0.60 & \RankOne{0.95} & \RankThree{4.4} \\
        S4Fusion~\cite{ma2025s4fusion}~{\venuesize \textcolor{venuedark}{[TIP'25]}} & \RankFour{6.88} & \RankTwo{37.70} & \RankThree{14.78} & 4.96 & 1.13 & 0.43 & \RankOne{0.93} & {6.7} & 6.57 & 31.97 & \RankFour{14.47} & 4.31 & 1.07 & 0.52 & \RankTwo{0.95} & 9.0 \\
        DCEvo~\cite{liu2025dcevo}~{\venuesize \textcolor{venuedark}{[CVPR'25]}} & 6.84 & 35.59 & 14.11 & 4.61 & 1.49 & 0.50 & \RankFour{0.81} & {7.0} & 6.67 & 34.75 & 13.77 & 4.04 & \RankThree{1.50} & 0.61 & \RankFour{0.89} & \RankFour{5.7} \\
        CIS-Fuse (Ours)~{\venuesize \textcolor{venuedark}{[Ours]}} & \RankOne{6.99} & \RankThree{37.35} & 14.66 & \RankThree{5.08} & \RankThree{1.71} & \RankThree{0.55} & 0.72 & \RankOne{3.6} & \RankOne{6.92} & \RankTwo{38.37} & \RankOne{14.67} & \RankTwo{4.54} & \RankTwo{1.70} & \RankFour{0.64} & 0.78 & \RankOne{2.7} \\
        \bottomrule
    \end{tabular}
    
    \\
    {\footnotesize
    \vspace{1mm}
    \begin{tabular}{@{}lcccc@{}}
        * The best, second-best, third-best, and fourth-best results are marked as &
        \RankOne{1st} & \RankTwo{2nd} & \RankThree{3rd} & \RankFour{4th}
    \end{tabular}
    }
    
    \end{tabular}
\vspace*{-14pt}
\end{table*}

\subsection{Training Strategy}
\label{sec:training}

\textbf{Output-mean supervision.}
As described in Sec.~\ref{sec:overall}, CIS-Fuse unrolls over $T$ time steps and produces a per-step output at each step. A common practice in SNN training is to apply the loss at every step independently and average the resulting losses, which we refer to as \emph{per-step-loss} supervision. We instead adopt \emph{output-mean} supervision: the per-step outputs are first averaged into the final fused image $\mathbf{I}_{\text{fused}}$, and the loss is then computed once on this aggregated output:
\begin{align}
    &\mathbf{I}_{\text{fused}} = \frac{1}{T}\sum_{t=1}^{T}\mathcal{N}(\mathbf{I}_{\text{ir}},\mathbf{I}_{\text{vis}};t), \\
    &\mathcal{L}_{\text{out-mean}} = \mathcal{L}\!\left(\mathbf{I}_{\text{fused}},\, \mathbf{I}_{\text{ir}},\, \mathbf{I}_{\text{vis}}\right),
\end{align}
where $\mathcal{N}(\cdot;t)$ denotes the network output at time step $t$. 
Although the per-step and output-mean losses appear similar, they are not equivalent for nonlinear losses, since $\mathcal{L}\!\left(\tfrac{1}{T}\sum_t \hat{\mathbf{I}}_t\right) \neq \tfrac{1}{T}\sum_t \mathcal{L}(\hat{\mathbf{I}}_t)$ in general. Output-mean loss aligns the training objective with the test-time aggregated output, whereas per-step-loss supervision biases the network toward strong single-step rather than strong aggregated predictions. Our ablation in Table~\ref{tab:abl_training} confirms that output-mean supervision yields better fusion quality.
% consistently better fusion quality across metrics.

\textbf{Loss function.}
Our overall training loss combines three terms: a maximum-intensity term for pixel-level fidelity, a maximum-gradient term for structural boundaries, and a sum-of-correlations-of-differences (SCD) term that encourages complementary information transfer from both modalities:
\begin{equation}
    \mathcal{L} = \mathcal{L}_{\text{int}} + \lambda_{\text{grad}}\,\mathcal{L}_{\text{grad}} + \lambda_{\text{scd}}\,\mathcal{L}_{\text{scd}}.
\end{equation}
The intensity term $\mathcal{L}_{\text{int}}$ pulls the fused image toward the element-wise maximum of the two sources, preserving both salient thermal targets and bright textured regions:
\begin{equation}
    \mathcal{L}_{\text{int}} = \big\|\max(\mathbf{I}_{\text{ir}}, \mathbf{I}_{\text{vis}}) - \mathbf{I}_{\text{fused}}\big\|_1.
\end{equation}
The gradient term $\mathcal{L}_{\text{grad}}$ applies the same maximum-selection principle in the gradient domain, sharpening object boundaries and fine textures:
\begin{equation}
    \mathcal{L}_{\text{grad}} = \big\|\max(|\nabla \mathbf{I}_{\text{ir}}|, |\nabla \mathbf{I}_{\text{vis}}|) - |\nabla \mathbf{I}_{\text{fused}}|\big\|_1,
\end{equation}
where $\nabla$ denotes the Sobel operator. The \emph{SCD} term~\cite{aslantas2015new} provides a global complementarity criterion that the two $\ell_1$ terms cannot capture on their own:
\begin{equation}
    \mathcal{L}_{\text{scd}} = -\bigl(\rho(\mathbf{I}_{\text{vis}}, \mathbf{I}_{\text{fused}} - \mathbf{I}_{\text{ir}}) + \rho(\mathbf{I}_{\text{ir}}, \mathbf{I}_{\text{fused}} - \mathbf{I}_{\text{vis}})\bigr),
\end{equation}
where $\rho(\cdot,\cdot)$ denotes the Pearson correlation coefficient. Intuitively, $\mathbf{I}_{\text{fused}} - \mathbf{I}_{\text{ir}}$ should resemble $\mathbf{I}_{\text{vis}}$ and vice versa, indicating that the fused image absorbs distinctive content from both sources. 
% The weighting coefficients $\{\lambda_{\text{grad}}, \lambda_{\text{scd}}\}$ are specified in Sec.~\ref{sec:implementation}.

\section{Experiments}
\subsection{Implementation Details}
\label{sec:implementation}

\textbf{Datasets.}
We train CIS-Fuse on the MSRS~\cite{tang2022piafusion} training set, which contains 1083 aligned infrared-visible image pairs. The trained model is then tested without any fine-tuning on four widely-used IVIF benchmarks, namely RoadScene~\cite{xu2020fusiondn}, MSRS~\cite{tang2022piafusion}, M3FD~\cite{liu2022target}, and FMB~\cite{liu2023multi}, with 50, 361, 300, and 280 test pairs, respectively. All images are converted to the YCbCr color space~\cite{li2025ustc,li2025loop}: only the Y channel is fused, and the visible Cb/Cr channels are recombined with the fused Y at inference to produce the color output.
% , and only the Y channel is fed into the network; the Cb and Cr channels of the visible image are kept and recombined with the fused Y channel at inference to produce the final color output.

\begin{figure*}[t!]
    \centering
    \includegraphics[width=.95\linewidth]{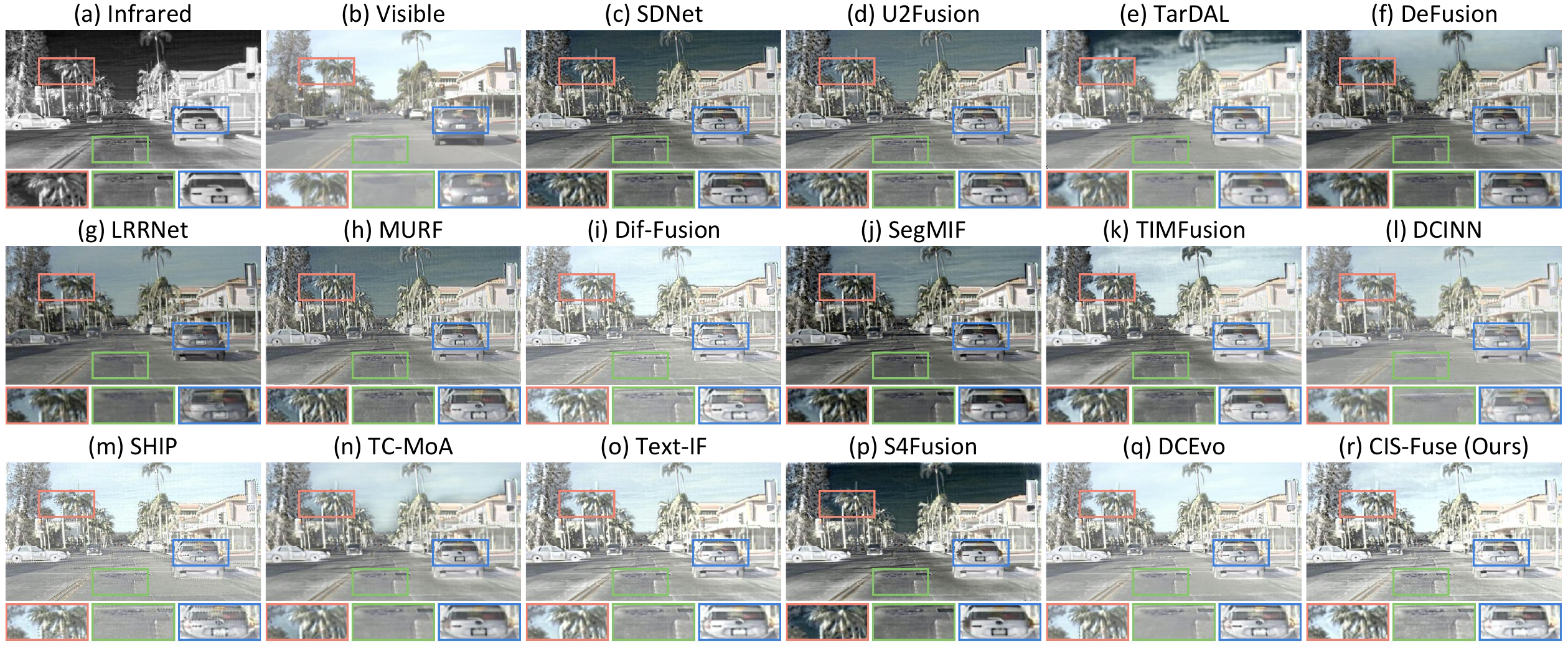}
    \vspace*{-8pt}
    \caption{Qualitative comparison on the RoadScene dataset. CIS-Fuse preserves salient thermal targets while retaining sharper textures and contrast than competing methods. Zoomed-in patches are shown at the bottom. Please enlarge for details.}
    \vspace*{-8pt}
    \label{fig:vis_roadscene}
\end{figure*}

\begin{figure*}[t!]
    \centering
    \includegraphics[width=.95\linewidth]{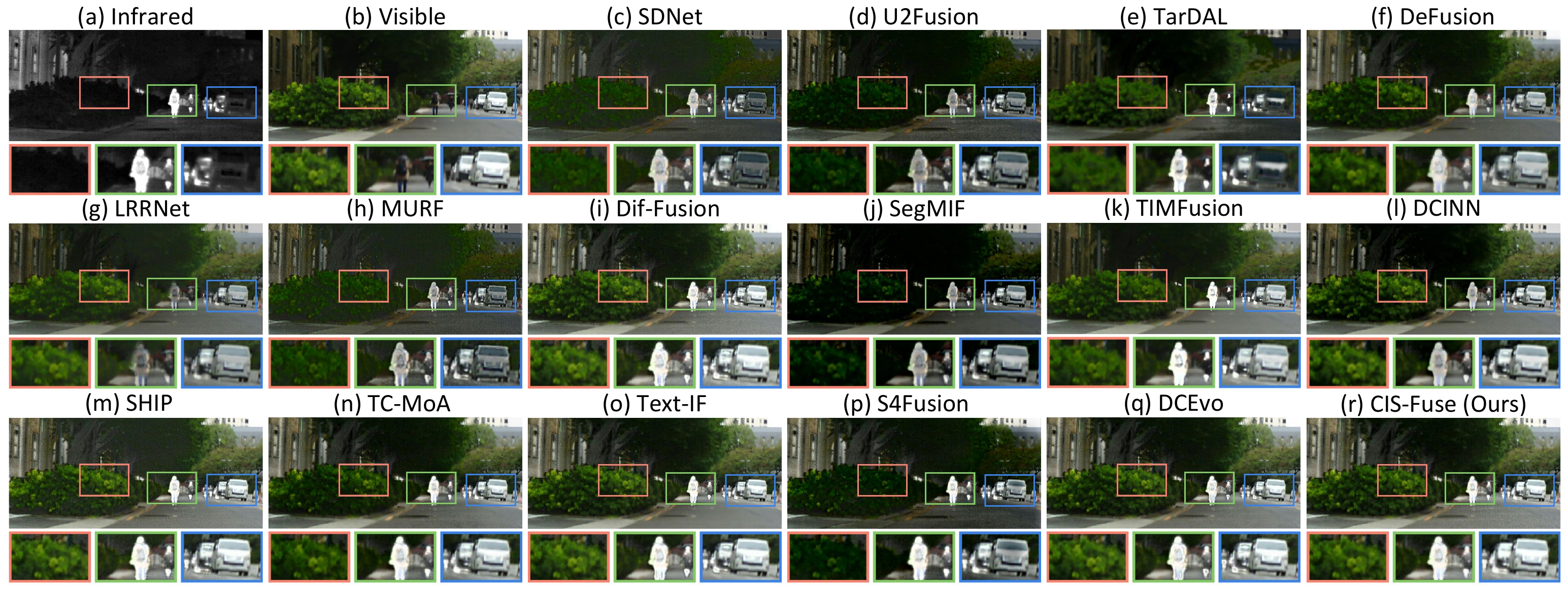}
    \vspace*{-8pt}
    \caption{Qualitative comparison on the MSRS dataset. CIS-Fuse retains strong thermal cues while recovering visible-side details that others over-smooth. Zoomed-in patches are shown at the bottom. Please enlarge for details.}
    \vspace*{-12pt}
    \label{fig:vis_msrs}
\end{figure*}

\textbf{Training settings.}
The network is trained end-to-end for 60 epochs with a batch size of 8 on an NVIDIA RTX 4090. We use the Adam optimizer with an initial learning rate of $3\times 10^{-4}$, which is halved at epochs 20 and 40. During training, each image pair is randomly cropped to $64\times 64$ patches. The number of spiking time steps is set to $T{=}4$, and all spiking neurons are PLIF with initial $\tau{=}2.0$ and a soft-reset scheme. The loss weights are set to $\lambda_{\text{grad}}{=}1.0$ and $\lambda_{\text{scd}}{=}0.1$.

\textbf{Architecture configuration.}
Both the IR and VIS encoders share the same architecture and parameters, with a feature dimension of $C{=}64$ and 3 ResBlockSNN blocks. The shallow BCMF branch uses $K{=}1$ block with the injection strength initialized to $\alpha_0{=}0.25$, while the deep BCMF branch uses $K{=}3$ blocks with $\alpha_0{=}0.75$. The RepHead uses parallel 3$\times$3, 1$\times$1, and 5$\times$5 branches during training, and is reparameterized into a single 5$\times$5 convolution at inference.

\subsection{Comparison on IVIF}
\label{sec:comparison_ivif}

\textbf{Compared methods.}

We compare CIS-Fuse against 15 representative state-of-the-art IVIF methods covering CNN-based, Transformer-based, generative, and task-driven paradigms, namely SDNet~\cite{zhang2021sdnet}, U2Fusion~\cite{xu2020u2fusion}, TarDAL~\cite{liu2022target}, DeFusion~\cite{liang2022fusion}, LRRNet~\cite{li2023lrrnet}, MURF~\cite{xu2023murf}, Dif-Fusion~\cite{yue2023dif}, SegMIF~\cite{liu2023multi}, TIMFusion~\cite{liu2024task}, DCINN~\cite{wang2024general}, SHIP~\cite{zheng2024probing}, TC-MoA~\cite{zhu2024task}, Text-IF~\cite{yi2024text}, S4Fusion~\cite{ma2025s4fusion}, and DCEvo~\cite{liu2025dcevo}. All competing methods are evaluated using their official pretrained models and default settings.

\textbf{Evaluation metrics.}
Following standard IVIF evaluation protocol, we adopt seven widely-used metrics: entropy (EN), standard deviation (SD), spatial frequency (SF), average gradient (AG), sum of correlations of differences (SCD)~\cite{aslantas2015new}, correlation coefficient (CC), and visual information fidelity for fusion (VIFF)~\cite{han2013new}. Higher values indicate better fusion quality for all seven metrics. To summarize cross-metric performance, we additionally report \emph{AvgRank}, the average ranking of each method across all seven metrics on a given benchmark, where a lower AvgRank indicates better performance.

\textbf{Quantitative results.}
Table~\ref{tab:quantitative_comparison} reports the quantitative comparison on RoadScene, MSRS, M3FD, and FMB. CIS-Fuse achieves the best AvgRank on all four benchmarks, consistently outperforming the strongest competitors such as Text-IF, DCEvo, and S4Fusion. In particular, CIS-Fuse ranks among the top methods on SD across the four benchmarks, indicating that the fused images preserve a wide intensity range and retain strong contrast from both modalities. 
% It also performs strongly on SCD, confirming that the cross-modal current injection in CIS effectively transfers complementary information from both sources, consistent with the motivation of operating at the membrane-potential level.

% \textbf{Quantitative results.}
% Table~\ref{tab:quantitative_comparison} reports the quantitative comparison on RoadScene, MSRS, M3FD, and FMB. CIS-Fuse achieves the best AvgRank on all four benchmarks, with values of 2.7, 3.1, 3.6, and 2.7, respectively, outperforming the strongest competitors Text-IF, DCEvo, and S4Fusion. In particular, CIS-Fuse attains the highest SD on three of the four datasets, reaching 53.32 on RoadScene, 43.70 on MSRS, and 38.37 on FMB, indicating that the fused images preserve a wide intensity range and retain strong contrast from both modalities. CIS-Fuse also leads in SCD on both RoadScene and MSRS, with scores of 1.79 and 1.86, respectively, confirming that the cross-modal current injection in CIS effectively transfers complementary information from both sources, consistent with the motivation of operating at the membrane-potential level. 
% % On VIFF, CIS-Fuse is competitive but does not always lead — this metric favors methods that aggressively boost visible-side textures, while CIS-Fuse balances both modalities; the corresponding qualitative results discussed below show that this balance translates into more faithful fused images.

\begin{figure*}
    \centering
    \includegraphics[width=.95\linewidth]{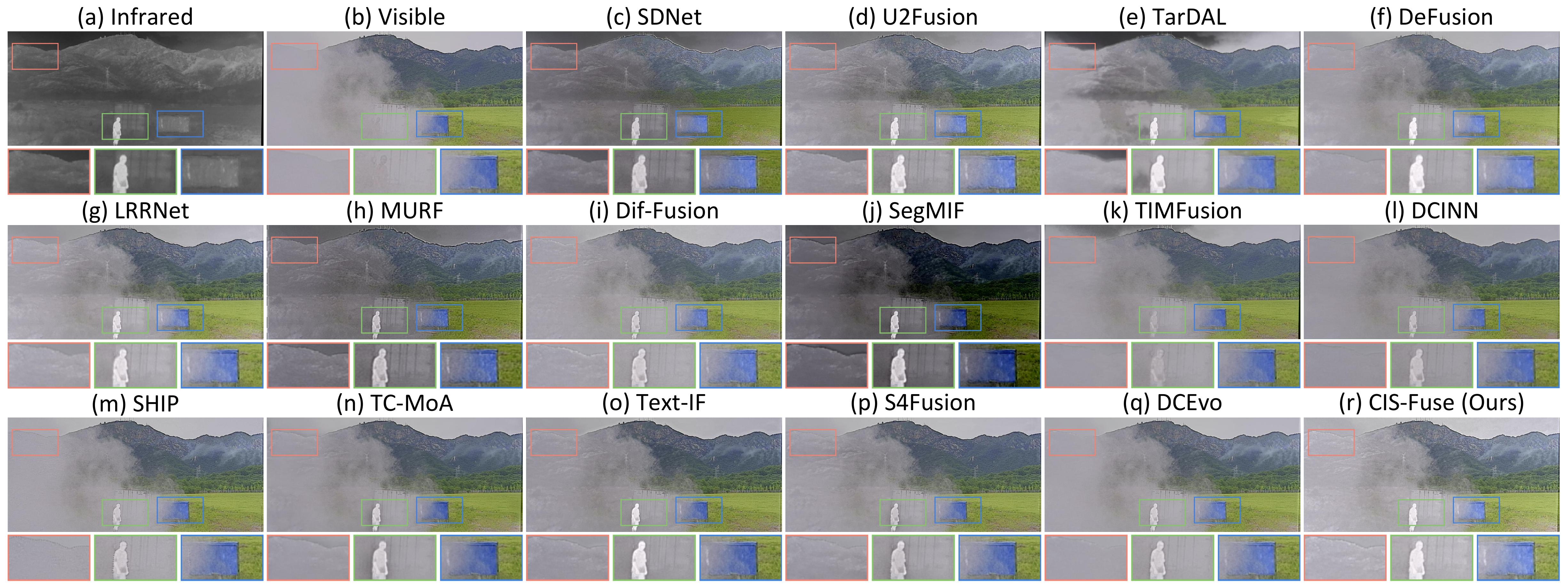}
    \vspace*{-8pt}
    \caption{Qualitative comparison on the M3FD dataset. 
    CIS-Fuse jointly preserves infrared targets and visible-side textures. In the zoomed-in patches at the bottom, the pedestrian stays sharp against the haze while the background retains fine texture. Please enlarge for details.}
    % \rrvs{CIS-Fuse jointly preserves infrared targets and visible-side textures. In the zoomed-in patches, the pedestrian stays sharp against the haze while the background retains fine texture.}
    % Zoomed-in patches are shown at the bottom. Please enlarge for details.}
    \vspace*{-8pt}
    \label{fig:vis_m3fd}
\end{figure*}

\begin{figure*}
    \centering
    \includegraphics[width=.95\linewidth]{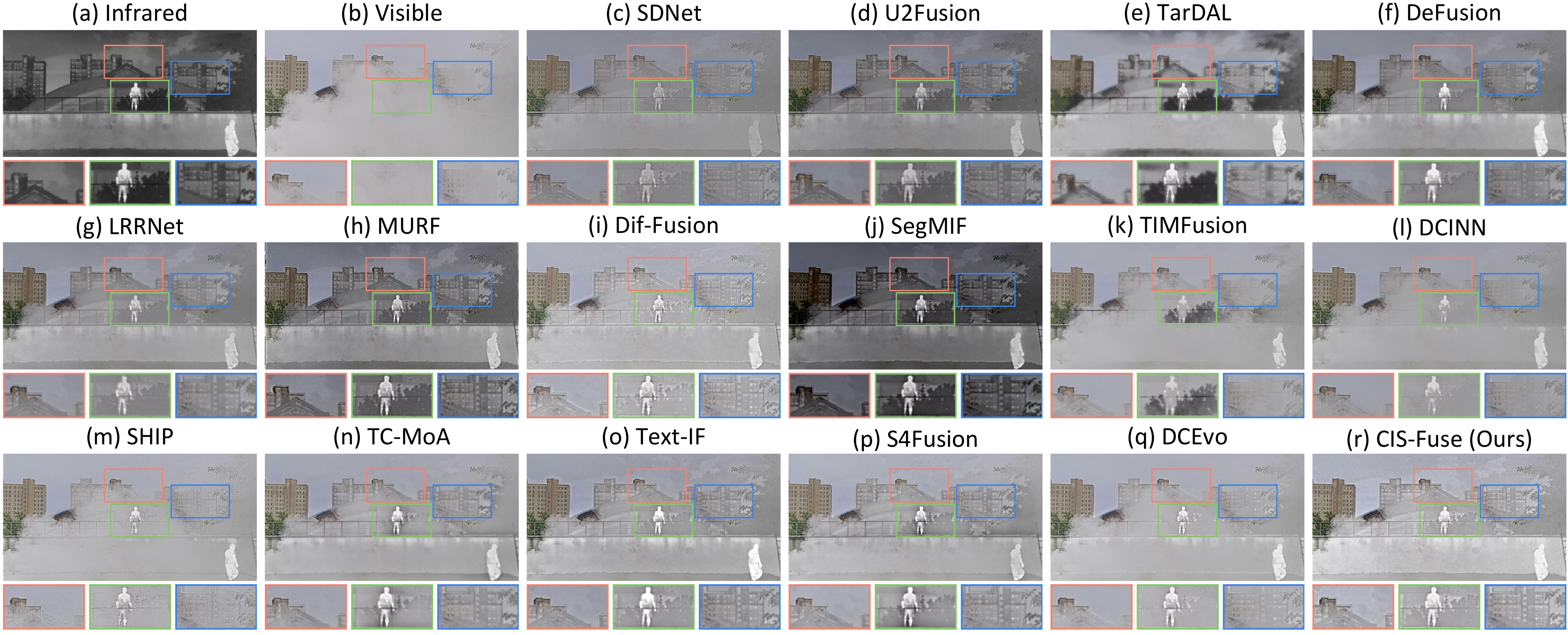}
    \vspace*{-8pt}
    \caption{Qualitative comparison on the FMB dataset. CIS-Fuse produces clearer object contours and richer scene context, especially in regions with strongly complementary modalities. Zoomed-in patches are shown at the bottom. Please enlarge for details.}
    \vspace*{-8pt}
    \label{fig:vis_fmb}
\end{figure*}

\begin{table}[t]
\centering
\renewcommand{\arraystretch}{1}
\setlength\tabcolsep{3.5pt}
\caption{Ablation studies on CIS-Fuse. Each variant alters only one design
dimension while keeping all other components identical to Ours.}
\label{tab:abl_structure}
\vspace*{-6pt}
\footnotesize
\begin{tabular}{lccccccc}
\toprule
\makecell[c]{Variant} & EN  & SD  & SF  & AG  & SCD  & CC  & VIFF  \\
\midrule
\multicolumn{8}{c}{{Group 1: Fusion mechanism}} \\
\midrule
Element-wise add                    & 6.60  & 40.89  & 10.88  & 3.65  & 1.80  & 0.62  & 0.89  \\
Concat + 1$\times$1 conv            & 6.61  & 41.73  & 10.49  & 3.48  & 1.82  & 0.63  & 0.85  \\
\rowcolor{gray!12}
Ours                                & 6.69  & 43.70  & 11.30  & 3.74  & 1.86  & 0.62  & 0.90  \\
\midrule
\multicolumn{8}{c}{{Group 2: BCMF directionality}} \\
\midrule
IR $\rightarrow$ VIS only           & 6.59  & 41.05  & 10.56  & 3.50  & 1.80  & 0.62  & 0.88  \\
VIS $\rightarrow$ IR only           & 6.62  & 40.70  & 10.70  & 3.58  & 1.79  & 0.63  & 0.90  \\
Non-alternating BCMF       & 6.70  & 41.81  & 10.80  & 3.58  & 1.82  & 0.62  & 0.86  \\
\rowcolor{gray!12}
Ours                                & 6.69  & 43.70  & 11.30  & 3.74  & 1.86  & 0.62  & 0.90  \\
\midrule
\multicolumn{8}{c}{{Group 3: Dual-branch deployment}} \\
\midrule
Symmetric ($K{=}2$)        & 6.66  & 41.92  & 10.75  & 3.57  & 1.82  & 0.62  & 0.90  \\
Inverted depth        & 6.63  & 41.97  & 10.81  & 3.58  & 1.82  & 0.62  & 0.89  \\
\rowcolor{gray!12}
Ours                                & 6.69  & 43.70  & 11.30  & 3.74  & 1.86  & 0.62  & 0.90  \\
\midrule
\multicolumn{8}{c}{{Group 4: Spiking neuron type}} \\
\midrule
IF neuron                           & 6.54  & 42.53  & 11.22  & 3.68  & 1.77  & 0.62  & 0.90  \\
LIF neuron                          & 6.55  & 41.48  & 10.86  & 3.58  & 1.78  & 0.62  & 0.85  \\
\rowcolor{gray!12}
Ours                                & 6.69  & 43.70  & 11.30  & 3.74  & 1.86  & 0.62  & 0.90  \\
\bottomrule
\end{tabular}
\vspace*{-12pt}
\end{table}

\textbf{Qualitative results.}
Fig.~\ref{fig:vis_roadscene}--\ref{fig:vis_fmb} show qualitative comparisons on the four benchmarks. As shown in Fig.~\ref{fig:vis_roadscene} for RoadScene, CIS-Fuse simultaneously preserves the sharp thermal silhouettes of vehicles and pedestrians and the fine textures of trees, road markings, and building façades, whereas several competing methods either dim the infrared targets, as seen in LRRNet and DCINN, or wash out visible-side textures, as seen in TarDAL. Fig.~\ref{fig:vis_msrs} presents results on MSRS, which features challenging low-light nighttime scenes; CIS-Fuse retains bright thermal cues such as pedestrians and headlights without overexposing them, and recovers visible-side details that other methods over-smooth. In Fig.~\ref{fig:vis_m3fd} for M3FD, CIS-Fuse produces more natural-looking fused images with clearer object contours under haze and complex outdoor illumination. Finally, Fig.~\ref{fig:vis_fmb} shows the results on FMB, where CIS-Fuse maintains clear scene structure and rich texture even when the two modalities provide highly complementary information.
Overall, these results show that membrane-potential-level fusion attains fusion quality comparable to ANN-based methods under the spike-based paradigm.

% Overall, the quantitative and qualitative results jointly demonstrate that performing cross-modal fusion at the membrane-potential level enables CIS-Fuse to achieve comparable fusion quality with ANN-based IVIF methods, while operating under the spike-based computation paradigm.

\begin{table}[t]
\centering
\renewcommand{\arraystretch}{1}
\setlength\tabcolsep{4pt}
\caption{Training strategy ablation studies. ``Output mean'' supervision and $T{=}4$
are adopted in our final model.}
\vspace*{-4pt}
\label{tab:abl_training}
\footnotesize
\begin{tabular}{lccccccc}
\toprule
Variant & EN  & SD  & SF  & AG  & SCD  & CC  & VIFF  \\
\midrule
\multicolumn{8}{c}{{Loss supervision strategy}} \\
\midrule
Last-step only            & 6.56  & 42.83  & 11.11  & 3.48  & 1.81  & 0.63  & 0.85  \\
Per-step loss             & 6.62  & 42.36  & 11.06  & 3.61  & 1.85  & 0.63  & 0.90  \\
\rowcolor{gray!12}
Output mean (Ours)        & 6.69  & 43.70  & 11.30  & 3.74  & 1.86  & 0.62  & 0.90  \\
\midrule
\multicolumn{8}{c}{{Number of spiking time steps $T$}} \\
\midrule
$T{=}1$                  & 6.69  & 42.48  & 11.05  & 3.71  & 1.85  & 0.63  & 0.87  \\
$T{=}2$                  & 6.65  & 42.66  & 11.32  & 3.74  & 1.83  & 0.63  & 0.87  \\
\rowcolor{gray!12}
$T{=}4$ (Ours)           & 6.69  & 43.70  & 11.30  & 3.74  & 1.86  & 0.62  & 0.90  \\
$T{=}8$                  & 6.70  & 42.13  & 11.15  & 3.74  & 1.82  & 0.62  & 0.91  \\
\bottomrule
\end{tabular}
\vspace*{-8pt}
\end{table}

\subsection{Ablation Studies}
\label{sec:ablation}
To validate the design choices of CIS-Fuse, we conduct ablation studies on the MSRS test set. Table~\ref{tab:abl_structure} examines four architectural design dimensions, where each variant alters only one component while keeping all other settings identical to the full model. Table~\ref{tab:abl_training} further examines the training strategy along two axes, namely the loss supervision scheme and the number of spiking time steps. We discuss each group in turn. As SNN-based IVIF baselines are scarce, all ablations below are conducted within an identical SNN framework, isolating the gain of CIS from that of the SNN paradigm itself.

\textbf{Fusion mechanism.}
We first compare our CIS-based fusion against two common baselines, element-wise addition and concatenation followed by a $1{\times}1$ convolution. As shown in Group 1 of Table~\ref{tab:abl_structure}, both baselines lag behind CIS on nearly every metric, with notable drops on SD and SCD. These results suggest that integrating the two modalities through gated current injection before spike firing is more effective than simple feature-level addition or concatenation.
% This indicates that simply mixing the two modalities at the feature level is insufficient. performing fusion at the membrane-potential level via gated current injection is the key to extracting complementary information from both sources.

\textbf{BCMF directionality.}
We then ablate the bidirectional and alternating-refinement design of BCMF. Unidirectional variants, namely IR$\rightarrow$VIS only and VIS$\rightarrow$IR only, consistently degrade fusion quality, confirming that neither modality alone should be treated as the sole modulator. Removing the alternating mechanism while keeping both directions, denoted Non-alternating BCMF, also underperforms our full design on SD and SF, showing that conditioning each update on the just-updated representation of the other modality is essential for effective mutual refinement.

\textbf{Dual-branch deployment.}
Group 3 of Table~\ref{tab:abl_structure} compares our asymmetric shallow-deep deployment against two alternatives. Setting both branches to identical depths $K{=}2$, denoted Symmetric, removes the soft inductive bias that drives functional specialization, while swapping the depths, denoted Inverted, contradicts the bias by pairing a strong injection strength with shallow stacking and vice versa. Both variants underperform the proposed shallow-$K{=}1$ plus deep-$K{=}3$ design, validating that the asymmetric configuration is what enables the functional divergence observed in Fig.~\ref{fig:alpha_dist}.

\textbf{Spiking neuron type.}
We further replace the PLIF neuron with the simpler IF and LIF neurons throughout the network. As shown in Group 4 of Table~\ref{tab:abl_structure}, both variants produce noticeably weaker fusion quality, particularly on EN, SD, and SCD. This confirms that the learnable membrane time constant in PLIF allows each spiking layer to adapt its temporal dynamics to the data, which is beneficial for the multi-step membrane-potential integration central to CIS-Fuse.

\textbf{Training strategy.}
Finally, Table~\ref{tab:abl_training} ablates the training strategy. For loss supervision, the last-step-only and per-step-loss variants both underperform the proposed output-mean supervision, confirming that aligning the training objective with the test-time aggregated output is important for nonlinear fusion losses such as SCD. For the number of spiking time steps, performance increases from $T{=}1$ to $T{=}4$ but saturates beyond that, with $T{=}8$ offering no consistent gain over $T{=}4$. We therefore adopt $T{=}4$ as a favorable trade-off between fusion quality and inference cost.

\renewcommand{\arraystretch}{1}
\setlength\tabcolsep{4.5pt}
% Table generated by Excel2LaTeX from sheet 'mAP50-95'
\begin{table}[t!]
  \centering
  \caption{Quantitative comparison of object detection on the M3FD dataset. Each fusion method is used to train a separate YOLOv8s detector under identical settings, and we report the per-class AP@[0.5:0.95] as well as the overall mAP@[0.5:0.95] averaged across all six categories.}
  \vspace*{-4pt}
\begin{tabular}{@{}l@{}}
    \begin{tabular}{lccccccc}
    \toprule
    \multicolumn{1}{c}{Method} & People & Car   & Bus   & Motor. & Lamp  & Truck & All \\
    \midrule
    Infrared                        & 55.9  & 68.3  & 73.9  & 42.2  & 41.3  & 55.3  & 56.1  \\
    Visible                         & 44.2  & 71.0  & 77.3  & 50.6  & 45.5  & 56.3  & 57.5  \\
    SDNet~\cite{zhang2021sdnet}     & 55.8  & 71.6  & 74.4  & 49.6  & 46.9  & 56.8  & \RankTwo{59.2}  \\
    U2Fusion~\cite{xu2020u2fusion}  & 54.8  & 71.3  & 74.5  & 50.9  & 45.4  & 57.4  & \RankThree{59.0}  \\
    TarDAL~\cite{liu2022target}     & 54.2  & 70.5  & 75.5  & 47.4  & 44.3  & 55.6  & 57.9  \\
    DeFusion~\cite{liang2022fusion} & 54.8  & 71.3  & 73.5  & 50.0  & 41.3  & 56.8  & 58.0  \\
    LRRNet~\cite{li2023lrrnet}      & 53.2  & 71.6  & 72.3  & 50.8  & 42.4  & 56.3  & 57.8  \\
    MURF~\cite{xu2023murf}          & 54.0  & 69.4  & 74.2  & 46.1  & 43.6  & 56.6  & 57.3  \\
    Dif-Fusion~\cite{yue2023dif}    & 54.8  & 71.2  & 77.1  & 50.2  & 42.6  & 56.8  & \RankFour{58.8}  \\
    SegMIF~\cite{liu2023multi}      & 54.3  & 71.1  & 74.9  & 48.7  & 47.5  & 55.0  & 58.6  \\
    TIMFusion~\cite{liu2024task}    & 52.0  & 70.8  & 73.2  & 48.7  & 43.1  & 58.3  & 57.7  \\
    DCINN~\cite{wang2024general}    & 54.2  & 70.2  & 72.3  & 47.7  & 40.7  & 62.1  & 57.9  \\
    SHIP~\cite{zheng2024probing}    & 54.3  & 70.8  & 75.5  & 49.8  & 42.6  & 55.1  & 58.0  \\
    TC-MoA~\cite{zhu2024task}       & 54.1  & 71.2  & 76.2  & 49.6  & 42.8  & 55.9  & 58.3  \\
    Text-IF~\cite{yi2024text}       & 54.4  & 71.4  & 72.6  & 50.1  & 44.7  & 57.8  & 58.5  \\
    S4Fusion~\cite{ma2025s4fusion}  & 53.7  & 71.6  & 73.4  & 51.3  & 42.4  & 55.9  & 58.0  \\
    DCEvo~\cite{liu2025dcevo}       & 54.8  & 71.0  & 73.7  & 50.2  & 43.0  & 58.3  & 58.5  \\
    CIS-Fuse (Ours)                 & 54.7  & 70.9  & 76.5  & 49.4  & 48.2  & 58.8  & \RankOne{59.8}  \\
    \bottomrule
    \end{tabular}%
    \\
    {\footnotesize
    \vspace{1mm}
    \begin{tabular}{@{}lccccccc}
        * 1st, 2nd, 3rd, 4th best: &
        \RankOne{1st} & \RankTwo{2nd} & \RankThree{3rd} & \RankFour{4th}
    \end{tabular}
    }
\end{tabular}
  \label{tab:det}%
  \vspace*{-18pt}
\end{table}%

\renewcommand{\arraystretch}{1}
\setlength\tabcolsep{4pt}
\begin{table*}[htbp]
  \centering
  \footnotesize
  \caption{Quantitative comparison of semantic segmentation on the MSRS dataset. Each fusion method trains a separate SegFormer segmenter under identical settings, and we report per-class IoU and pixel accuracy together with the overall mIoU and mAcc. The first two rows report single-modality references.}
  \vspace*{-4pt}
\begin{tabular}{@{}l@{}}
    \begin{tabular}{lcccccccccccccccccccc}
    \toprule
    \multirow{2}{*}{Method} & \multicolumn{2}{c}{unlabelled} & \multicolumn{2}{c}{car} & \multicolumn{2}{c}{person} & \multicolumn{2}{c}{bike} & \multicolumn{2}{c}{curve} & \multicolumn{2}{c}{car\_stop} & \multicolumn{2}{c}{guardrail} & \multicolumn{2}{c}{color\_cone} & \multicolumn{2}{c}{bump} & \multirow{2}{*}{mIoU} & \multirow{2}{*}{mAcc} \\
    \cmidrule(lr){2-3}  \cmidrule(lr){4-5}  \cmidrule(lr){6-7}  \cmidrule(lr){8-9}  \cmidrule(lr){10-11}  \cmidrule(lr){12-13}  \cmidrule(lr){14-15}  \cmidrule(lr){16-17}  \cmidrule(lr){18-19}
     & IoU  & Acc  & IoU  & Acc  & IoU  & Acc  & IoU  & Acc  & IoU  & Acc  & IoU  & Acc  & IoU  & Acc  & IoU  & Acc  & IoU  & Acc \\
    \midrule
    Infrared    & 98.2  & 99.3  & 87.1  & 92.2  & 71.3  & 83.6  & 68.8  & 77.3  & 56.7  & 64.6  & 64.8  & 75.6  & 55.1  & 57.7  & 54.7  & 63.4  & 72.2  & 78.6  & 69.9  & 76.9  \\
    Visible    & 98.2  & 99.3  & 89.1  & 93.7  & 61.6  & 73.0  & 69.6  & 79.0  & 58.0  & 68.3  & 69.5  & 80.2  & 70.0  & 74.6  & 59.4  & 72.6  & 73.2  & 80.2  & 72.1  & 80.1  \\
    SDNet~\cite{zhang2021sdnet} & 98.3  & 99.4  & 88.3  & 92.7  & 72.5  & 84.0  & 70.0  & 80.3  & 58.6  & 67.3  & 67.7  & 77.0  & 71.5  & 79.0  & 55.6  & 63.8  & 71.9  & 80.0  & 72.7  & 80.4  \\
    U2Fusion~\cite{xu2020u2fusion} & 98.4  & 99.4  & 88.6  & 92.8  & 71.2  & 82.9  & 70.4  & 79.8  & 59.2  & 67.8  & 69.2  & 78.4  & 76.0  & 81.5  & 57.5  & 65.8  & 73.9  & 81.5  & \RankThree{73.8}  & 81.1  \\
    TarDAL~\cite{liu2022target} & 98.1  & 99.3  & 87.5  & 92.0  & 66.6  & 78.7  & 67.1  & 75.9  & 52.8  & 60.6  & 66.0  & 75.9  & 74.4  & 79.1  & 50.1  & 58.7  & 73.1  & 80.6  & 70.6  & 77.9  \\
    DeFusion~\cite{liang2022fusion} & 98.3  & 99.4  & 88.4  & 92.6  & 69.9  & 82.0  & 69.9  & 78.9  & 59.0  & 68.6  & 67.4  & 78.2  & 69.3  & 73.8  & 58.7  & 67.2  & 75.9  & 83.1  & 73.0  & 80.4  \\
    LRRNet~\cite{li2023lrrnet} & 98.4  & 99.4  & 89.4  & 93.7  & 71.4  & 83.4  & 70.2  & 79.1  & 59.5  & 70.1  & 70.9  & 79.9  & 72.6  & 77.9  & 58.5  & 67.4  & 74.0  & 80.4  & \RankTwo{73.9}  & \RankThree{81.3}  \\
    MURF~\cite{xu2023murf}  & 98.3  & 99.3  & 88.0  & 92.3  & 71.4  & 83.0  & 69.7  & 79.5  & 59.0  & 67.4  & 66.1  & 76.2  & 66.7  & 70.3  & 55.6  & 64.2  & 71.6  & 80.3  & 71.8  & 79.2  \\
    Dif-Fusion~\cite{yue2023dif} & 98.4  & 99.4  & 89.0  & 93.6  & 71.1  & 82.8  & 70.0  & 79.1  & 59.4  & 69.2  & 69.4  & 78.7  & 71.6  & 76.5  & 57.1  & 66.4  & 71.7  & 78.3  & 73.1  & 80.4  \\
    SegMIF~\cite{liu2023multi} & 98.3  & 99.4  & 88.0  & 92.4  & 71.0  & 83.2  & 70.7  & 79.6  & 59.0  & 67.9  & 69.5  & 79.9  & 73.0  & 76.6  & 55.2  & 65.7  & 77.4  & 83.7  & 73.6  & 80.9  \\
    TIMFusion~\cite{liu2024task} & 98.2  & 99.3  & 87.1  & 91.6  & 67.1  & 78.6  & 70.3  & 79.3  & 58.2  & 67.8  & 69.4  & 79.2  & 76.0  & 81.9  & 54.3  & 66.2  & 71.0  & 78.4  & 72.4  & 80.3  \\
    DCINN~\cite{wang2024general} & 98.3  & 99.4  & 88.2  & 92.4  & 70.9  & 82.7  & 69.4  & 78.6  & 59.2  & 68.1  & 69.9  & 80.4  & 70.1  & 74.4  & 54.1  & 65.3  & 73.4  & 80.7  & 72.6  & 80.2  \\
    SHIP~\cite{zheng2024probing}  & 98.4  & 99.4  & 88.4  & 93.0  & 71.3  & 83.3  & 69.9  & 78.6  & 59.4  & 69.0  & 71.4  & 81.0  & 68.1  & 71.7  & 57.4  & 66.0  & 74.3  & 80.9  & 73.2  & 80.3  \\
    TC-MoA~\cite{zhu2024task} & 98.3  & 99.4  & 87.9  & 92.4  & 70.4  & 82.3  & 69.9  & 79.0  & 58.0  & 67.9  & 69.6  & 79.1  & 68.0  & 72.8  & 57.8  & 66.7  & 76.4  & 83.8  & 72.9  & 80.4  \\
    Text-IF~\cite{yi2024text} & 98.3  & 99.3  & 87.9  & 92.9  & 70.3  & 82.1  & 69.5  & 78.8  & 59.5  & 69.6  & 71.0  & 79.8  & 73.6  & 78.5  & 55.1  & 66.5  & 71.0  & 77.4  & 72.9  & 80.5  \\
    S4Fusion~\cite{ma2025s4fusion} & 98.3  & 99.4  & 87.5  & 91.8  & 70.5  & 82.0  & 70.5  & 79.0  & 59.9  & 69.7  & 68.1  & 78.2  & 76.3  & 81.0  & 57.7  & 68.1  & 74.1  & 81.7  & 73.7  & \RankFour{81.2}  \\
    DCEvo~\cite{liu2025dcevo} & 98.4  & 99.4  & 88.9  & 93.4  & 70.9  & 82.5  & 70.5  & 79.8  & 60.0  & 69.8  & 72.4  & 80.9  & 71.0  & 75.8  & 55.9  & 66.4  & 76.1  & 83.4  & \RankFour{73.8}  & \RankTwo{81.3}  \\
    CIS-Fuse (Ours)  & 98.4  & 99.3  & 88.6  & 93.4  & 71.2  & 83.1  & 70.2  & 79.6  & 59.4  & 69.2  & 72.2  & 82.4  & 74.7  & 79.9  & 58.6  & 66.6  & 75.3  & 82.2  & \RankOne{74.3}  & \RankOne{81.7}  \\
    \bottomrule
    \end{tabular}%
    \\
    {\footnotesize
    \vspace{1mm}
    \begin{tabular}{@{}lccccccc}
        * 1st, 2nd, 3rd, 4th best: &
        \RankOne{1st} & \RankTwo{2nd} & \RankThree{3rd} & \RankFour{4th}
    \end{tabular}
    }
\end{tabular}

  \label{tab:seg}%
  \vspace*{-12pt}
\end{table*}%

\subsection{Comparison on Downstream Tasks}
\label{sec:downstream}

Beyond visual quality, a key motivation for IVIF is to provide more informative inputs for downstream perception. We evaluate CIS-Fuse on two representative tasks, object detection on M3FD and semantic segmentation on MSRS, and compare against the same 15 IVIF baselines used in Sec.~\ref{sec:comparison_ivif}.

\textbf{Object detection.}
We adopt YOLOv8s via Ultralytics as the detector, initialized from its official COCO-pretrained weights. For each fusion method, we fine-tune a separate detector on its fused images using the official M3FD train split, and then evaluate on the M3FD test split. All detectors are trained for 200 epochs with a batch size of 16 and an initial learning rate of 0.01, at an input resolution of $640\times 640$. We report per-class average precision (AP) at IoU thresholds from 0.5 to 0.95 with a step of 0.05, denoted AP@[0.5:0.95], as well as the overall mAP averaged over all classes.

As shown in Table~\ref{tab:det}, CIS-Fuse achieves the best overall mAP of 59.8, surpassing the strongest baselines SDNet and U2Fusion. The per-class behavior suggests that the fused images simultaneously preserve thermal saliency for compact targets and visible-side texture for larger structured objects.

\textbf{Semantic segmentation.}
We adopt SegFormer~\cite{xie2021segformer} with a MiT-B2 backbone as the segmenter, where the backbone is initialized from ImageNet-pretrained weights and the segmentation head is randomly initialized for the 9 MSRS classes. Following the same protocol as in detection, a separate segmenter is fine-tuned on the fused images of each method using the MSRS standard train split and evaluated on the standard test split. All segmenters are trained for 100 epochs with a batch size of 4 and an initial learning rate of $6\times 10^{-5}$, at an input resolution of $512\times 512$. We report per-class IoU and pixel accuracy across the nine MSRS categories, together with the overall mIoU and mAcc.

\begin{figure*}
    \centering
    \includegraphics[width=.95\linewidth]{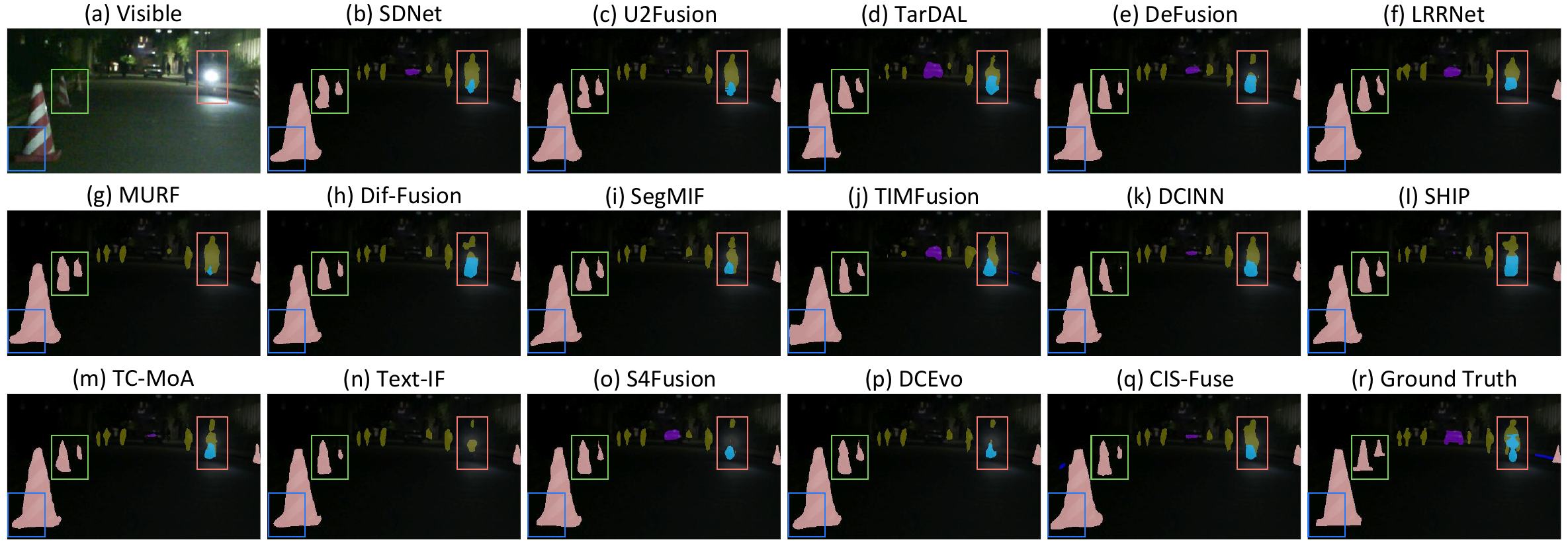}
    \vspace*{-10pt}
    \caption{Qualitative comparison of semantic segmentation on the MSRS dataset. Predicted masks are overlaid on the visible image.}
    \label{fig:vis_seg}
    \vspace*{-8pt}
\end{figure*}

As shown in Table~\ref{tab:seg}, CIS-Fuse achieves the best overall mIoU of 74.3 and mAcc of 81.7, and remains competitive on safety-critical categories such as car\_stop, guardrail, and color\_cone, where reliable segmentation benefits from both thermal cues and visible-side texture. Fig.~\ref{fig:vis_seg} further provides qualitative comparisons, where CIS-Fuse yields cleaner object boundaries and more complete masks on small safety-critical structures than competing methods. The consistent gains on both detection and segmentation indicate that the membrane-potential-level cross-modal integration in CIS-Fuse transfers well to high-level perception tasks.

\subsection{Computational Costs}

We analyze the computational cost of CIS-Fuse by measuring its spike firing rates, from which the \textit{theoretical} inference energy is then estimated under a standard operation-level cost model and compared against recent ANN-based methods.

\begin{figure}[t]
    \centering
    \includegraphics[width=.95\linewidth]{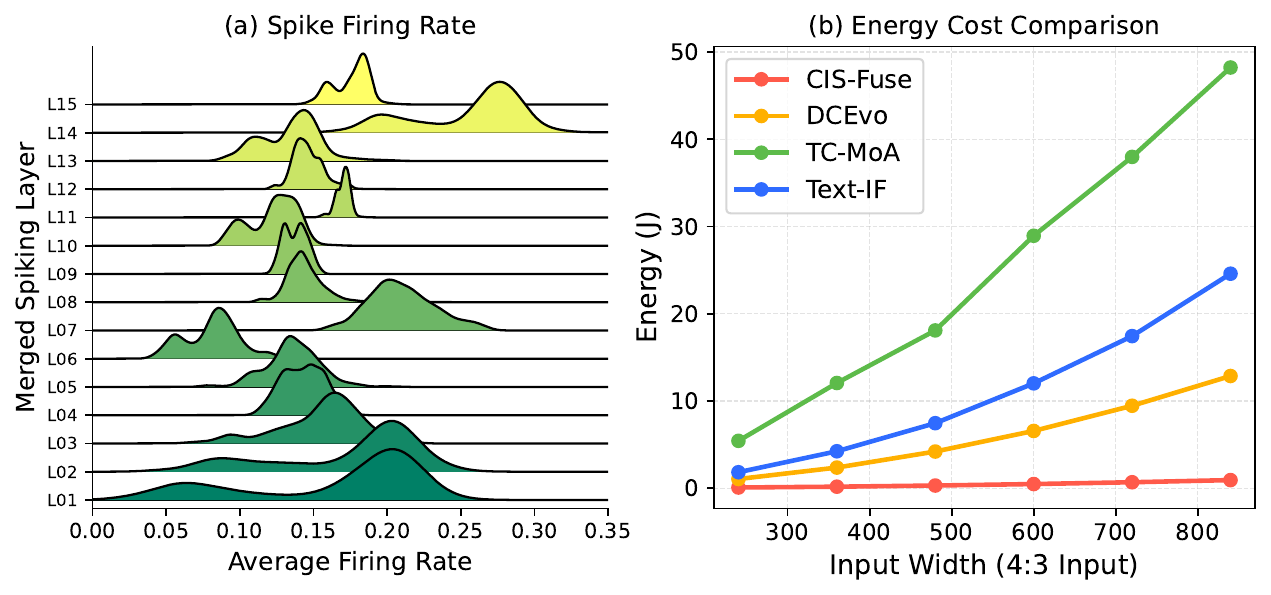}
    \vspace*{-10pt}
    \caption{(a) Per-sample average firing rates of the 15 merged spiking layers of CIS-Fuse on the M3FD test set, with paired sn1/sn2 averaged per sample. L01--L03 correspond to the encoder, L04--L06 to the shallow BCMF branch, L07--L13 to the deep BCMF branch, and L14--L15 to the decoder. (b) Total inference energy versus input width at a 4:3 aspect ratio, compared with three recent ANN-based methods.}
    \label{fig:energy}
    \vspace*{-14pt}
\end{figure}

\textbf{Spike firing rate.}
Fig.~\ref{fig:energy} (a) shows the per-sample distributions of the average firing rate at each spiking layer of CIS-Fuse on the M3FD dataset~\cite{liu2022target}. For layers containing a paired $(\mathrm{sn}_1, \mathrm{sn}_2)$ within the same residual unit, the two are averaged per sample to form a single merged layer, yielding 15 merged layers in total. The 15 layers cover the full forward path, where L01--L03 correspond to the encoder, L04--L06 to the shallow BCMF branch, L07--L13 to the deep BCMF branch, and L14--L15 to the decoder. The network-wide average firing rate is $r \approx 14.48\%$, indicating that the majority of spike-driven operations reduce to additions skipped at zero input.

% \textbf{Energy model.}
% Following the convention used in prior SNN works~\cite{zhu2022event}, we estimate inference energy by assigning a fixed energy cost to each multiply-accumulate (MAC) in ANN layers and each accumulate (AC) in SNN layers:
\textbf{Energy model.}
Following the prior SNN literature~\cite{zhu2022event}, we provide a \textit{theoretical} energy estimate by assigning a fixed energy cost to each multiply-accumulate (MAC) in ANN layers and each accumulate (AC) in SNN layers:
\begin{equation}
E_{\mathrm{ANN}} = N_{\mathrm{MAC}}^{\mathrm{ANN}} \cdot e_{\mathrm{MAC}}, \quad
E_{\mathrm{SNN}} = N_{\mathrm{AC}}^{\mathrm{SNN}} \cdot e_{\mathrm{AC}},
\end{equation}
where we adopt the widely used $e_{\mathrm{MAC}} = 4.6$\,pJ and $e_{\mathrm{AC}} = 0.9$\,pJ. Unlike ANN MACs that are triggered by every input activation, an SNN AC is triggered only when the presynaptic neuron emits a spike. Its effective operation count is therefore scaled by the layer-wise firing rate:
\begin{equation}
N_{\mathrm{AC}}^{\mathrm{SNN}} = T \cdot \sum_{l} r_l \cdot N_{\mathrm{MAC},l}^{\mathrm{raw}},
\end{equation}
where $N_{\mathrm{MAC},l}^{\mathrm{raw}}$ is the dense MAC count of the $l$-th spiking layer if it were treated as an ANN layer, $r_l$ is its measured average firing rate, and $T$ is the number of spiking time steps. For CIS-Fuse, we use the per-layer firing rates measured on M3FD with $T = 4$.

\rrvs{These values follow the 45\,nm CMOS measurements of Horowitz~\cite{horowitz20141} and are widely adopted in the SNN energy-efficiency literature~\cite{zhu2022event,yao2023spike}. The same model is applied to both ANNs and SNNs, counting dense MACs for the former and firing-rate-scaled ACs for the latter, giving a first-order architectural estimate rather than an on-device measurement.}

\noindent\textbf{Energy comparison.}
Using the above model, we compare CIS-Fuse against three recent ANN-based methods, DCEvo~\cite{liu2025dcevo}, TC-MoA~\cite{zhu2024task}, and Text-IF~\cite{yi2024text}. These four methods have 1.54M (deployed), 2.01M, 340.58M, and 215.12M parameters, respectively. We vary the input width from 320 to 800 under a fixed 4:3 aspect ratio. As shown in Fig.~\ref{fig:energy} (b), CIS-Fuse consistently incurs a lower energy cost across the range, with the gap widening at higher resolutions. 
Beyond fusion quality, the result also indicates that the energy efficiency intrinsic to spike-based computation is well preserved under our membrane-potential-level fusion design.
% Although low power is not the primary goal of this work, the result indicates that the energy efficiency intrinsic to spike-based computation is well preserved under our membrane-potential-level fusion design.

\section{Conclusion}
In this paper, we explored infrared and visible image fusion under the spiking neural network paradigm. To preserve fine-grained cross-modal responses that may otherwise be lost through binary spikes, we proposed to perform fusion at the continuous membrane-potential level before firing. We developed the Current Injection Spiking (CIS) operator, which injects a gated auxiliary current into the main-modality neuron, and extended it into a Bidirectional Cross-Modal Fusion (BCMF) block with alternating refinement. Combined with an asymmetric dual-branch architecture, a reparameterized output head, and output-mean supervision, CIS-Fuse achieves competitive fusion quality on four IVIF benchmarks and downstream perception tasks. Energy analysis based on the measured firing rates further suggests that CIS-Fuse requires less inference energy than representative ANN-based methods.

\bibliographystyle{IEEEtran}
\bibliography{ref}

@String(TPAMI  = {IEEE TPAMI})

@String(IJCV  = {IJCV})

@String(TIP   = {IEEE TIP})

@String(TCSVT = {IEEE TCSVT})

@String(TMM   =	{IEEE TMM})

@String(CVPR  = {CVPR})

@String(ICCV  = {ICCV})

@String(ECCV  = {ECCV})

@String(ICML  = {ICML})

@String(NeurIPS  = {NeurIPS})

@String(IJCAI = {IJCAI})

@String(AAAI = {AAAI})

@String(IROS = {IROS})

@article{toet1989merging,
  title={Merging thermal and visual images by a contrast pyramid},
  author={Toet, Alexander and Van Ruyven, Lodewik J and Valeton, J Mathee},
  journal={Optical Engineering},
  volume={28},
  number={7},
  pages={789--792},
  year={1989},
}

@inproceedings{ha2017mfnet,
  title={{MFNet}: Towards real-time semantic segmentation for autonomous vehicles with multi-spectral scenes},
  author={Ha, Qishen and Watanabe, Kohei and Karasawa, Takumi and Ushiku, Yoshitaka and Harada, Tatsuya},
  booktitle=IROS,
  pages={5108--5115},
  year={2017},
}

@inproceedings{li2018cross,
  title={Cross-modal ranking with soft consistency and noisy labels for robust {RGB-T} tracking},
  author={Li, Chenglong and Zhu, Chengli and Huang, Yan and Tang, Jin and Wang, Liang},
  booktitle=ECCV,
  pages={808--823},
  year={2018}
}

@article{li2024object,
  title={Object segmentation-assisted inter prediction for versatile video coding},
  author={Li, Zhuoyuan and Yuan, Zikun and Li, Li and Liu, Dong and Tang, Xiaohu and Wu, Feng},
  journal={IEEE TBC},
  volume={70},
  number={4},
  pages={1236--1253},
  year={2024},
}

@article{tang2022image,
  title={Image fusion in the loop of high-level vision tasks: A semantic-aware real-time infrared and visible image fusion network},
  author={Tang, Linfeng and Yuan, Jiteng and Ma, Jiayi},
  journal={Information Fusion},
  volume={82},
  pages={28--42},
  year={2022},
}

@article{xu2020u2fusion,
  title={{U2Fusion}: A unified unsupervised image fusion network},
  author={Xu, Han and Ma, Jiayi and Jiang, Junjun and Guo, Xiaojie and Ling, Haibin},
  journal=TPAMI,
  volume={44},
  number={1},
  pages={502--518},
  year={2020},
}

@article{zhang2021sdnet,
  title={{SDNet}: A versatile squeeze-and-decomposition network for real-time image fusion},
  author={Zhang, Hao and Ma, Jiayi},
  journal=IJCV,
  volume={129},
  number={10},
  pages={2761--2785},
  year={2021},
  publisher={Springer}
}

@inproceedings{zhao2023cddfuse,
  title={{CDDFuse}: Correlation-driven dual-branch feature decomposition for multi-modality image fusion},
  author={Zhao, Zixiang and Bai, Haowen and Zhang, Jiangshe and Zhang, Yulun and Xu, Shuang and Lin, Zudi and Timofte, Radu and Van Gool, Luc},
  booktitle=CVPR,
  pages={5906--5916},
  year={2023}
}

@inproceedings{yi2024text,
  title={{Text-IF}: Leveraging semantic text guidance for degradation-aware and interactive image fusion},
  author={Yi, Xunpeng and Xu, Han and Zhang, Hao and Tang, Linfeng and Ma, Jiayi},
  booktitle=CVPR,
  pages={27026--27035},
  year={2024}
}

@inproceedings{zhu2024task,
  title={Task-customized mixture of adapters for general image fusion},
  author={Zhu, Pengfei and Sun, Yang and Cao, Bing and Hu, Qinghua},
  booktitle=CVPR,
  pages={7099--7108},
  year={2024}
}

@article{ma2019fusiongan,
  title={{FusionGAN}: A generative adversarial network for infrared and visible image fusion},
  author={Ma, Jiayi and Yu, Wei and Liang, Pengwei and Li, Chang and Jiang, Junjun},
  journal={Information Fusion},
  volume={48},
  pages={11--26},
  year={2019},
  publisher={Elsevier}
}

@inproceedings{wang2022unsupervised,
  title={Unsupervised misaligned infrared and visible image fusion via cross-modality image generation and registration},
  author={Wang, Di and Liu, Jinyuan and Fan, Xin and Liu, Risheng},
  booktitle=IJCAI,
  pages={3508--3515},
  year={2022}
}

@inproceedings{zhao2023ddfm,
  title={{DDFM}: Denoising diffusion model for multi-modality image fusion},
  author={Zhao, Zixiang and Bai, Haowen and Zhu, Yuanzhi and Zhang, Jiangshe and Xu, Shuang and Zhang, Yulun and Zhang, Kai and Meng, Deyu and Timofte, Radu and Van Gool, Luc},
  booktitle=ICCV,
  pages={8082--8093},
  year={2023}
}

@article{yue2023dif,
  title={{Dif-Fusion}: Toward high color fidelity in infrared and visible image fusion with diffusion models},
  author={Yue, Jun and Fang, Leyuan and Xia, Shaobo and Deng, Yue and Ma, Jiayi},
  journal=TIP,
  volume={32},
  pages={5705--5720},
  year={2023},
  publisher={IEEE}
}

@inproceedings{liu2022target,
  title={Target-aware dual adversarial learning and a multi-scenario multi-modality benchmark to fuse infrared and visible for object detection},
  author={Liu, Jinyuan and Fan, Xin and Huang, Zhanbo and Wu, Guanyao and Liu, Risheng and Zhong, Wei and Luo, Zhongxuan},
  booktitle=CVPR,
  pages={5802--5811},
  year={2022}
}

@inproceedings{liu2023multi,
  title={Multi-interactive feature learning and a full-time multi-modality benchmark for image fusion and segmentation},
  author={Liu, Jinyuan and Liu, Zhu and Wu, Guanyao and Ma, Long and Liu, Risheng and Zhong, Wei and Luo, Zhongxuan and Fan, Xin},
  booktitle=ICCV,
  pages={8115--8124},
  year={2023}
}

@inproceedings{liu2025dcevo,
  title={{DCEvo}: Discriminative cross-dimensional evolutionary learning for infrared and visible image fusion},
  author={Liu, Jinyuan and Zhang, Bowei and Mei, Qingyun and Li, Xingyuan and Zou, Yang and Jiang, Zhiying and Ma, Long and Liu, Risheng and Fan, Xin},
  booktitle=CVPR,
  pages={2226--2235},
  year={2025}
}

@article{maass1997networks,
  title={Networks of spiking neurons: the third generation of neural network models},
  author={Maass, Wolfgang},
  journal={Neural Networks},
  volume={10},
  number={9},
  pages={1659--1671},
  year={1997},
}

@article{tavanaei2019deep,
  title={Deep learning in spiking neural networks},
  author={Tavanaei, Amirhossein and Ghodrati, Masoud and Kheradpisheh, Saeed Reza and Masquelier, Timoth{\'e}e and Maida, Anthony},
  journal={Neural Networks},
  volume={111},
  pages={47--63},
  year={2019},
}

@article{roy2019towards,
  title={Towards spike-based machine intelligence with neuromorphic computing},
  author={Roy, Kaushik and Jaiswal, Akhilesh and Panda, Priyadarshini},
  journal={Nature},
  volume={575},
  number={7784},
  pages={607--617},
  year={2019},
}

@article{london2005dendritic,
  title={Dendritic computation},
  author={London, Michael and H{\"a}usser, Michael},
  journal={Annual Review of Neuroscience},
  volume={28},
  number={1},
  pages={503--532},
  year={2005},
}

@inproceedings{xu2020fusiondn,
  title={{FusionDN}: A unified densely connected network for image fusion},
  author={Xu, Han and Ma, Jiayi and Le, Zhuliang and Jiang, Junjun and Guo, Xiaojie},
  booktitle=AAAI,
  volume={34},
  number={07},
  pages={12484--12491},
  year={2020}
}

@article{tang2022piafusion,
  title={{PIAFusion}: A progressive infrared and visible image fusion network based on illumination aware},
  author={Tang, Linfeng and Yuan, Jiteng and Zhang, Hao and Jiang, Xingyu and Ma, Jiayi},
  journal={Information Fusion},
  volume={83},
  pages={79--92},
  year={2022},
}

@article{livingstone1988segregation,
  title={Segregation of form, color, movement, and depth: anatomy, physiology, and perception},
  author={Livingstone, Margaret and Hubel, David},
  journal={Science},
  volume={240},
  number={4853},
  pages={740--749},
  year={1988},
}

@inproceedings{liang2022fusion,
  title={Fusion from decomposition: A self-supervised decomposition approach for image fusion},
  author={Liang, Pengwei and Jiang, Junjun and Liu, Xianming and Ma, Jiayi},
  booktitle=ECCV,
  pages={719--735},
  year={2022},
}

@article{li2023lrrnet,
  title={{LRRNet}: A novel representation learning guided fusion network for infrared and visible images},
  author={Li, Hui and Xu, Tianyang and Wu, Xiao-Jun and Lu, Jiwen and Kittler, Josef},
  journal=TPAMI,
  volume={45},
  number={9},
  pages={11040--11052},
  year={2023},
}

@inproceedings{zheng2024probing,
  title={Probing synergistic high-order interaction in infrared and visible image fusion},
  author={Zheng, Naishan and Zhou, Man and Huang, Jie and Hou, Junming and Li, Haoying and Xu, Yuan and Zhao, Feng},
  booktitle=CVPR,
  pages={26384--26395},
  year={2024}
}

@article{xu2023murf,
  title={{MURF}: Mutually reinforcing multi-modal image registration and fusion},
  author={Xu, Han and Yuan, Jiteng and Ma, Jiayi},
  journal=TPAMI,
  volume={45},
  number={10},
  pages={12148--12166},
  year={2023},
  publisher={IEEE}
}

@inproceedings{zhao2024image,
  title={Image fusion via vision-language model},
  author={Zhao, Zixiang and Deng, Lilun and Bai, Haowen and Cui, Yukun and Zhang, Zhipeng and Zhang, Yulun and Qin, Haotong and Chen, Dongdong and Zhang, Jiangshe and Wang, Peng and others},
  booktitle=ICML,
  pages={60749--60765},
  year={2024}
}

@article{ma2025s4fusion,
  title={{S4Fusion}: Saliency-aware selective state space model for infrared and visible image fusion},
  author={Ma, Haolong and Li, Hui and Cheng, Chunyang and Wang, Gaoang and Song, Xiaoning and Wu, Xiao-Jun},
  journal=TIP,
  year={2025},
  publisher={IEEE}
}

@article{wang2024general,
  title={A general paradigm with detail-preserving conditional invertible network for image fusion},
  author={Wang, Wu and Deng, Liang-Jian and Ran, Ran and Vivone, Gemine},
  journal=IJCV,
  volume={132},
  number={4},
  pages={1029--1054},
  year={2024},
}

@inproceedings{zhao2023metafusion,
  title={{MetaFusion}: Infrared and visible image fusion via meta-feature embedding from object detection},
  author={Zhao, Wenda and Xie, Shigeng and Zhao, Fan and He, You and Lu, Huchuan},
  booktitle=CVPR,
  pages={13955--13965},
  year={2023}
}

@article{liu2024task,
  title={A task-guided, implicitly-searched and meta-initialized deep model for image fusion},
  author={Liu, Risheng and Liu, Zhu and Liu, Jinyuan and Fan, Xin and Luo, Zhongxuan},
  journal=TPAMI,
  volume={46},
  number={10},
  pages={6594--6609},
  year={2024},
  publisher={IEEE}
}

@article{neftci2019surrogate,
  title={Surrogate gradient learning in spiking neural networks},
  author={Neftci, Emre O and Mostafa, Hesham and Zenke, Friedemann},
  journal={IEEE SPM},
  volume={36},
  number={6},
  pages={51--63},
  year={2019},
}

@inproceedings{wu2019direct,
  title={Direct training for spiking neural networks: Faster, larger, better},
  author={Wu, Yujie and Deng, Lei and Li, Guoqi and Zhu, Jun and Xie, Yuan and Shi, Luping},
  booktitle=AAAI,
  volume={33},
  number={01},
  pages={1311--1318},
  year={2019}
}

@inproceedings{zheng2021going,
  title={Going deeper with directly-trained larger spiking neural networks},
  author={Zheng, Hanle and Wu, Yujie and Deng, Lei and Hu, Yifan and Li, Guoqi},
  booktitle=AAAI,
  volume={35},
  number={12},
  pages={11062--11070},
  year={2021}
}

@inproceedings{fang2021deep,
  title={Deep residual learning in spiking neural networks},
  author={Fang, Wei and Yu, Zhaofei and Chen, Yanqi and Huang, Tiejun and Masquelier, Timoth{\'e}e and Tian, Yonghong},
  booktitle=NeurIPS,
  volume={34},
  pages={21056--21069},
  year={2021}
}

@inproceedings{fang2021incorporating,
  title={Incorporating learnable membrane time constant to enhance learning of spiking neural networks},
  author={Fang, Wei and Yu, Zhaofei and Chen, Yanqi and Masquelier, Timoth{\'e}e and Huang, Tiejun and Tian, Yonghong},
  booktitle=ICCV,
  pages={2661--2671},
  year={2021}
}

@inproceedings{hagenaars2021self,
  title={Self-supervised learning of event-based optical flow with spiking neural networks},
  author={Hagenaars, Jesse and Paredes-Vall{\'e}s, Federico and De Croon, Guido},
  booktitle=NeurIPS,
  volume={34},
  pages={7167--7179},
  year={2021}
}

@inproceedings{li2025brain,
  title={Brain-inspired spiking neural networks for energy-efficient object detection},
  author={Li, Ziqi and Gao, Tao and An, Yisheng and Chen, Ting and Zhang, Jing and Wen, Yuanbo and Liu, Mengkun and Zhang, Qianxi},
  booktitle=CVPR,
  pages={3552--3562},
  year={2025}
}

@inproceedings{zhu2022event,
  title={Event-based video reconstruction via potential-assisted spiking neural network},
  author={Zhu, Lin and Wang, Xiao and Chang, Yi and Li, Jianing and Huang, Tiejun and Tian, Yonghong},
  booktitle=CVPR,
  pages={3594--3604},
  year={2022}
}

@article{xiao2026learning,
  title={Learning Dual Modality Interactions for Event-based Motion Deblurring},
  author={Xiao, Zeyu and Li, Zhuoyuan and Zhao, Yang and Liu, Yu and Zhang, Zhao and Jia, Wei},
  journal=TMM,
  year={2026},
  publisher={IEEE}
}

@inproceedings{xiao2025event,
  title={Event-based video super-resolution via state space models},
  author={Xiao, Zeyu and Wang, Xinchao},
  booktitle=CVPR,
  pages={12564--12574},
  year={2025}
}

@article{zhao2026spike,
  author={Zhao, Rui and Xiong, Ruiqin and Wang, Dongkai and Xuan, Shiyu and Zhang, Jian and Fan, Xiaopeng and Huang, Tiejun},
  journal=TPAMI, 
  title={Spike Camera Optical Flow Estimation Based on Continuous Spike Streams}, 
  year={2026},
  volume={48},
  number={4},
  pages={4756-4770},
}

@article{zheng2026spikecv,
  title={{SpikeCV}: open a continuous computer vision era},
  author={Zheng, Yajing and Zhang, Jiyuan and Zhao, Rui and Ding, Jianhao and Chen, Shiyan and Wu, Weijian and Xiong, Ruiqin and Yu, Zhaofei and Huang, Tiejun},
  journal={Science China Information Sciences},
  volume={69},
  number={3},
  pages={132106},
  year={2026},
}

@inproceedings{zhao2024boosting,
  title={Boosting spike camera image reconstruction from a perspective of dealing with spike fluctuations},
  author={Zhao, Rui and Xiong, Ruiqin and Zhao, Jing and Zhang, Jian and Fan, Xiaopeng and Yu, Zhaofei and Huang, Tiejun},
  booktitle=CVPR,
  pages={24955--24965},
  year={2024}
}

@inproceedings{zhao2022learning,
  title={Learning optical flow from continuous spike streams},
  author={Zhao, Rui and Xiong, Ruiqin and Zhao, Jing and Yu, Zhaofei and Fan, Xiaopeng and Huang, Tiejun},
  booktitle=NeurIPS,
  volume={35},
  pages={7905--7920},
  year={2022}
}

@article{zhao2023spike,
  title={Spike camera image reconstruction using deep spiking neural networks},
  author={Zhao, Rui and Xiong, Ruiqin and Zhang, Jian and Yu, Zhaofei and Zhu, Shuyuan and Ma, Lei and Huang, Tiejun},
  journal=TCSVT,
  volume={34},
  number={6},
  pages={5207--5212},
  year={2024},
  publisher={IEEE}
}

@inproceedings{ding2021repvgg,
  title={{RepVGG}: Making vgg-style convnets great again},
  author={Ding, Xiaohan and Zhang, Xiangyu and Ma, Ningning and Han, Jungong and Ding, Guiguang and Sun, Jian},
  booktitle=CVPR,
  pages={13733--13742},
  year={2021}
}

@article{aslantas2015new,
  title={A new image quality metric for image fusion: The sum of the correlations of differences},
  author={Aslantas, Veysel and Bendes, Emre},
  journal={AEU International Journal of Electronics and Communications},
  volume={69},
  number={12},
  pages={1890--1896},
  year={2015},
}

@article{han2013new,
  title={A new image fusion performance metric based on visual information fidelity},
  author={Han, Yu and Cai, Yunze and Cao, Yin and Xu, Xiaoming},
  journal={Information Fusion},
  volume={14},
  number={2},
  pages={127--135},
  year={2013},
  publisher={Elsevier}
}

@inproceedings{xie2021segformer,
  title={{SegFormer}: Simple and efficient design for semantic segmentation with transformers},
  author={Xie, Enze and Wang, Wenhai and Yu, Zhiding and Anandkumar, Anima and Alvarez, Jose M and Luo, Ping},
  booktitle=NeurIPS,
  pages={12077--12090},
  year={2021}
}

@article{li2025ustc,
  title={{USTC-TD}: A test dataset and benchmark for image and video coding in 2020s},
  author={Li, Zhuoyuan and Liao, Junqi and Tang, Chuanbo and Zhang, Haotian and Li, Yuqi and Bian, Yifan and Sheng, Xihua and Feng, Xinmin and Li, Yao and Gao, Changsheng and others},
  journal=TMM,
  year={2026},
  volume={28},
  number={},
  pages={269-284},
  publisher={IEEE}
}

@article{zhang2025spiking,
  title={Spiking neural networks with adaptive membrane time constant for event-based tracking},
  author={Zhang, Jiqing and Zhang, Malu and Wang, Yuanchen and Liu, Qianhui and Yin, Baocai and Li, Haizhou and Yang, Xin},
  journal=TIP,
  volume={34},
  pages={1009--1021},
  year={2025},
  publisher={IEEE}
}

@article{zhan2024spiking,
  title={Spiking transfer learning from rgb image to neuromorphic event stream},
  author={Zhan, Qiugang and Liu, Guisong and Xie, Xiurui and Tao, Ran and Zhang, Malu and Tang, Huajin},
  journal=TIP,
  volume={33},
  pages={4274--4287},
  year={2024},
  publisher={IEEE}
}

@article{li2024spiking,
  title={Spiking tucker fusion transformer for audio-visual zero-shot learning},
  author={Li, Wenrui and Wang, Penghong and Xiong, Ruiqin and Fan, Xiaopeng},
  journal=TIP,
  volume={33},
  pages={4840--4852},
  year={2024},
  publisher={IEEE}
}

@article{li2025spiking,
  title={Spiking variational graph representation inference for video summarization},
  author={Li, Wenrui and Han, Wei and Deng, Liang-Jian and Xiong, Ruiqin and Fan, Xiaopeng},
  journal=TIP,
  year={2025},
  publisher={IEEE}
}

@article{tan2024rle,
  title={{RLE}: A unified perspective of data augmentation for cross-spectral re-identification},
  author={Tan, Lei and Zhang, Yukang and Han, Keke and Dai, Pingyang and Zhang, Yan and Wu, Yongjian and Ji, Rongrong},
  journal=NeurIPS,
  volume={37},
  pages={126977--126996},
  year={2024}
}

@article{li2025multi,
  title={When Multi-Focus Image Fusion Meets Nonlinear Spiking Neural {P} Systems},
  author={Li, Bo and Zhang, Lingling and Bao, Tingting and Lei, Yunkuo and Zhang, Xiaoqing and Liu, Jun},
  journal=TMM,
  year={2025},
  publisher={IEEE}
}

@inproceedings{horowitz20141,
  title={1.1 computing's energy problem (and what we can do about it)},
  author={Horowitz, Mark},
  booktitle={ISSCC},
  pages={10--14},
  year={2014},
  organization={IEEE}
}

@inproceedings{yao2023spike,
  title={Spike-driven transformer},
  author={Yao, Man and Hu, Jiakui and Zhou, Zhaokun and Yuan, Li and Tian, Yonghong and Xu, Bo and Li, Guoqi},
  booktitle=NeurIPS,
  volume={36},
  pages={64043--64058},
  year={2023}
}

@inproceedings{luo2024integer,
  title={Integer-valued training and spike-driven inference spiking neural network for high-performance and energy-efficient object detection},
  author={Luo, Xinhao and Yao, Man and Chou, Yuhong and Xu, Bo and Li, Guoqi},
  booktitle=ECCV,
  pages={253--272},
  year={2024},
  organization={Springer}
}

@inproceedings{xiao2026spiking,
  title={Spiking meets attention: Efficient remote sensing image super-resolution with attention spiking neural networks},
  author={Xiao, Yi and Yuan, Qiangqiang and Jiang, Kui and Huang, Wenke and Zhang, Qiang and Zheng, Tingting and Lin, Chia-Wen and Zhang, Liangpei},
  booktitle=NeurIPS,
  volume={38},
  pages={65240--65259},
  year={2025}
}

@article{li2025loop,
  title={In-Loop Filtering Using Learned Look-Up Tables for Video Coding},
  author={Li, Zhuoyuan and Li, Jiacheng and Li, Yao and Li, Jialin and Li, Li and Liu, Dong and Wu, Feng},
  journal={arXiv preprint arXiv:2509.09494},
  year={2025}
}

@inproceedings{wang2025sample,
  title={SAMPLE: Semantic Alignment through Temporal-Adaptive Multimodal Prompt Learning for Event-Based Open-Vocabulary Action Recognition},
  author={Wang, Jing and Zhao, Rui and Xiong, Ruiqin and Wang, Xingtao and Fan, Xiaopeng and Huang, Tiejun},
  booktitle=ICCV,
  pages={14409--14419},
  year={2025}
}

@article{liu2024infrared,
  title={Infrared and visible image fusion: From data compatibility to task adaption},
  author={Liu, Jinyuan and Wu, Guanyao and Liu, Zhu and Wang, Di and Jiang, Zhiying and Ma, Long and Zhong, Wei and Fan, Xin and Liu, Risheng},
  journal=TPAMI,
  volume={47},
  number={4},
  pages={2349--2369},
  year={2024},
  publisher={IEEE}
}

\end{document}